\pdfoutput=1

\documentclass[11pt]{article}

\usepackage{acl}

\usepackage{times}
\usepackage{latexsym}

\usepackage[T1]{fontenc}

\usepackage[utf8]{inputenc}

\usepackage{microtype}

\usepackage{tabulary}
\usepackage{tabularx}
\usepackage{amsmath}
\usepackage{amssymb}
\usepackage{MnSymbol}%
\usepackage{wasysym}%
\usepackage{tikz}
\usepackage{pgfplots}
\usepgfplotslibrary{groupplots}
\usetikzlibrary{calc}
\usepackage{xcolor}
\usepackage{color, colortbl}
\usepackage[normalem]{ulem}
\usepackage{url}
\usepackage{multirow}
\usepackage{float}
\usepackage{paralist}
\usepackage{comment}
\usepackage{enumitem}

\pgfplotsset{compat=newest}

\usepackage{color}
\usepackage{colortbl,array,xcolor}
\usepackage{silence}

\usepackage{multicol}
\usepackage{multirow}
\usepackage{booktabs}
\usepackage{soul}
\usepackage{pgfplots}
\usepackage{dsfont}
\usepackage{bbm}
\usepackage{adjustbox}
\usepackage{tabu}
\usepackage{CJKutf8}

\usepgfplotslibrary{groupplots}
\usetikzlibrary{matrix}

\newcolumntype{?}{!{\vrule width 1pt}}

\usepackage{mathtools}
\usepackage{arydshln}

\usepackage{graphicx}  %
\graphicspath{{images/}{plots/}}
\usepackage{pgfplots}
\pgfplotsset{compat=1.12}

\usepackage{nccmath}
\usepackage{comment}

\definecolor{darkgreen}{HTML}{228B22}
\definecolor{cyellow}{HTML}{CC79A7}
\definecolor{nred}{HTML}{D55E00}
\definecolor{nblue}{HTML}{0072B2}
\definecolor{ngreen}{HTML}{009E73}
\definecolor{ggreen}{HTML}{82B366}
\definecolor{bblue}{HTML}{6C8EBF}

\usepackage{pifont}

\usepackage{xspace}

\newcommand{\interalia}[1]{\citep[\emph{inter alia}]{#1}}
\newcommand{\genie}{\textsc{Genie}\xspace}
\newcommand{\thumb}{\textsc{THumB}\xspace}

\usepackage[T1]{fontenc}

\usepackage[utf8]{inputenc}

\usepackage{microtype}

\usepackage{longtable}
\usepackage{verbatimbox}

\title{Transparent Human Evaluation for Image Captioning}

\author{
    Jungo Kasai$^{\heartsuit}$\thanks{\ \ Work was done during an internship at AI2.}  
\quad
\textbf{Keisuke Sakaguchi}$^{\clubsuit}$
\quad
\textbf{Lavinia Dunagan}$^{\heartsuit}$
\quad 
\textbf{Jacob Morrison}$^{\heartsuit\spadesuit}$
\\
\textbf{Ronan Le Bras}$^{\clubsuit}$
\quad
\textbf{Yejin Choi}$^{\heartsuit\clubsuit}$
\quad
\textbf{Noah A.\ Smith}$^{\heartsuit\clubsuit}$\\ \\
$^{\heartsuit}$Paul G.\ Allen School of Computer Science \& Engineering, University of Washington
    \\
     $^{\clubsuit}$Allen Institute for AI  \quad 
     $^{\spadesuit}$Department of Linguistics, University of Washington  \\
    {\tt \{jkasai,laviniad,jacobm00,yejin,nasmith\}@cs.washington.edu}\\
    {\tt \{keisukes,ronanlb\}@allenai.org}
}

\begin{document}
\maketitle
\begin{abstract}
We establish \thumb, a rubric-based human evaluation protocol for image captioning models.
Our scoring rubrics and their definitions are carefully developed based on machine- and human-generated captions on the MSCOCO dataset.
Each caption is evaluated along two main dimensions in a tradeoff (\textit{precision} and \textit{recall}) as well as other aspects that measure the text quality (\textit{fluency}, \textit{conciseness}, and \textit{inclusive language}).
Our evaluations demonstrate several critical problems of the current evaluation practice.
Human-generated captions show substantially higher quality than machine-generated ones, especially in coverage of salient information (i.e., recall), while most automatic metrics say the opposite.
Our rubric-based results reveal that CLIPScore, a recent metric that uses image features, better correlates with human judgments than conventional text-only metrics because it is more sensitive to recall.
We hope that this work will promote a more transparent evaluation protocol for image captioning and its automatic metrics.\footnote{All data are available at \url{https://github.com/jungokasai/THumB}.}
\end{abstract}

\section{Introduction}
Recent progress in large-scale training has pushed the state of the art in vision-language tasks \interalia{li2020oscar, zhang2021vinvl}.
One of these tasks is
image captioning, whose objective is to generate a caption that describes the given image.
The performance in image captioning has been primarily measured in automatic metrics (e.g., CIDEr, \citealp{cider2015}; SPICE, \citealp{SPICE16}) on popular benchmarks, such as MSCOCO \cite{mscoco} and Flickr8k \cite{flicker8k-expert-crowd}. %
Use of these metrics is justified based on their correlation with human judgments collected in previous work \interalia{flicker8k-expert-crowd, elliott-keller-2014-comparing, kilickaya-etal-2017-evaluating}.

\begin{figure}[t]
\centering
    \includegraphics[width=0.48\textwidth]{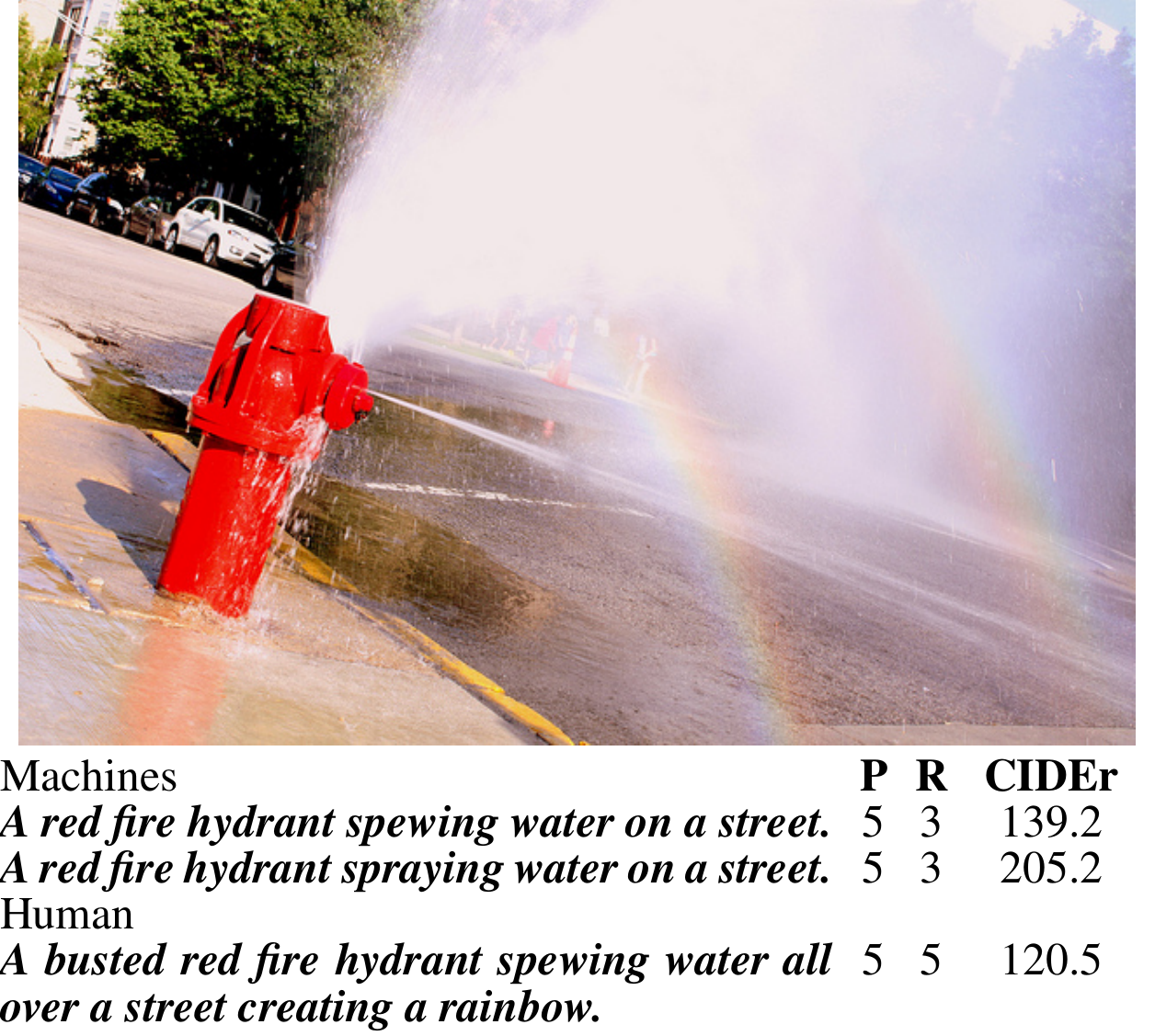}
\caption{These machine captions are \textit{precise} (in the scale of 1--5) but lose points in recall (i.e., coverage of salient information); they both ignore the rainbow in the picture. Automatic metrics, such as CIDEr, do not capture this failure.}
\label{fig:hydrant_rainbow}
\vspace{-0.4cm}
\end{figure}

Continuous use of these previous human judgments, however, raises significant concerns for development of both captioning models and automatic metrics because of their lack of transparency.
In previous work, annotators (crowdworkers, typically) rate image captions directly \cite{flicker8k-expert-crowd}, pairwise \cite{cider2015}, or along multiple dimensions such as thoroughness \cite{images-to-sentences15} and truthfulness \cite{yatskar-etal-2014-see}.
These scoring judgments depend highly on individual annotators' discretion and understanding of the annotation scheme \cite{freitag2021experts, clark-etal-2021-thats}, making it difficult to decompose, interpret, and validate annotations.
This lack of transparency also makes it difficult to interpret evaluation results for downstream applications where some aspects are particularly important (e.g., accessibility for people with visual impairments; \citealp{Gleason19, GleasonCHI20}).
Further, these annotations were done only on relatively old models (e.g., MSCOCO leaderboard submissions in 2015; \citealp{SPICE16}).
Correlations of automatic metrics with human judgments can break down especially when model types change \cite{callison-burch-etal-2006-evaluating}, or generation models become increasingly powerful \cite{ma-etal-2019-results, Edunov2020OnTE}.
We thus develop an up-to-date, transparent human evaluation protocol to better understand how current models perform and how automatic metrics are correlated when applied to current models.

At the core of our rubrics are two main scores in a tradeoff: \textit{precision} and \textit{recall} (Fig.\ \ref{fig:hydrant_rainbow}).
The former measures accuracy of the information in a caption, and the latter assesses how much of the salient information in the image is covered.
We then penalize a caption if we find a problem in \textit{fluency}, \textit{conciseness}, or \textit{inclusive language}.
Two or more authors evaluate every instance and collaborate to resolve disagreements, ensuring high quality of the annotations.
We assess outputs from four strong models as well as human-generated reference captions from MSCOCO.
We call our scores \thumb 1.0 (\textbf{T}ransparent \textbf{Hum}an \textbf{B}enchmark), and release them publicly.
\paragraph{Key Findings}
We made several key observations from the evaluations.
\begin{itemize}[noitemsep, leftmargin=*, topsep=1em]
  \setlength\itemsep{1em}
\item Machine-generated captions from recent models have been claimed to achieve superhuman performance using popular automatic metrics (human performance is ranked at the 250th place in the MSCOCO leaderboard),\footnote{\url{https://competitions.codalab.org/competitions/3221\#results}.} but they still show substantially lower quality than human-generated ones.
\item Machines fall short of humans, especially in recall (Fig.\ \ref{fig:hydrant_rainbow}), but most automatic metrics say the opposite.
This finding is consistent with prior work that showed that machines tend to produce less diverse captions than humans \cite{van-miltenburg-etal-2018-measuring}.
\item Human performance is underestimated in the current leaderboard paradigm, and there is still much room for improvement on MSCOCO captioning.
\item CLIPScore and RefCLIPScore \cite{clipscore}, recently proposed metrics that use image features, improve correlations particularly in recall.
While they fail to score human generation much higher than machine one, they capture an aspect that is less reflected in text-only metrics.
\item Currently available strong captioning models generate highly fluent captions. Fluency evaluation is thus no longer crucial in ranking these models.
\end{itemize}

\section{Evaluation Protocol}
We establish a transparent evaluation protocol for English image captioning models.
Our rubrics and rules are developed through discussions among all annotators (first four authors of this paper) and designed to increase the reliability of evaluation \cite{Jonsson2007TheUO}

\subsection{Evaluation Setups and Quality Control}
We used images from the test data in the standard \textit{Karpathy split} \cite{Karpathy_2015_CVPR} of the MSCOCO dataset \cite{mscoco}.
The dataset consists of 113K, 5K, and 5K train/dev./test everyday-scene photos sampled from Flickr.
We randomly sampled 500 test images and prepared one human- and four machine-generated captions for every image (\S\ref{sec:evaluated-captions}). 
We first performed developmental evaluations of 250 captions for 50 images and created rubrics.
We then proceeded with the rest of the captions.
For every image, captions were shuffled, and thus annotators did not know which caption corresponded to which model, thereby avoiding a potential bias from knowledge about the models.
We conducted two-stage annotations: the first annotator scores all captions for given images, and the second annotator checks and modifies the scores when necessary.
After the developmental phase, the $\kappa$ coefficient \cite{Cohen1960ACO} was 0.86 in precision and 0.82 in recall for the rest of the evaluated captions (\S\ref{sec:main_scores}).\footnote{Furthermore, we found that a third annotator did not change the scores for all 100 captions randomly sampled for meta-evaluations, confirming the sufficiently high quality of our two-stage annotations.
Disagreement in ratings can also result from a certain degree of subjectivity \cite{Misra2016SeeingTT}.
}
The first four authors of this paper conducted all evaluations; none of them are color blind or low vision, two are native English speakers, and one is a graduate student in linguistics.
We finally ensured that at least one native speaker evaluated the fluency of every caption (\S\ref{sec:penalties}), meaning that if a caption is annotated by the two non-native speakers, one native speaker checks the fluency in an additional round.

\subsection{\thumb 1.0}
Similar to the framework of the automatic SPICE metric \cite{SPICE16}, we base our manual evaluations on two main scores: \textbf{precision} and \textbf{recall}.
We also consider three types of penalty: \textbf{fluency}, \textbf{conciseness}, and \textbf{inclusive language}.
The overall score is computed by averaging precision and recall and deducting penalty points.

\begin{table*}[h!]
    \small
     \begin{center}
     \begin{tabular}{  @{} c l  c @{\hspace{0.2cm}} c  c  c @{\hspace{0.1cm}} c   @{} }
     \toprule
      Image & Caption & P & R & Flu.& Total\\ 
    \cmidrule(lr){1-1}\cmidrule(lr){2-2}
    \cmidrule(lr){3-5}
    \cmidrule(lr){6-6}
    \raisebox{0.0cm}{
     \multirow{6}{*}{
     {\includegraphics[width=0.21\textwidth]{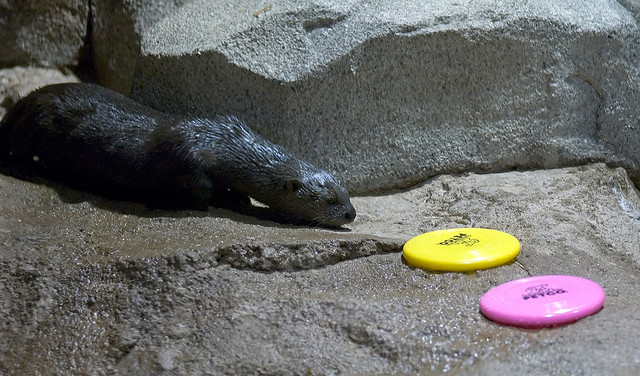}}
     }}
     & 
     \textbf{1-A}: Up-Down 
     &  
     \multirow{2}{*}{3}
     &
     \multirow{2}{*}{4}
     &
     \multirow{2}{*}{0}
     & 
     \multirow{2}{*}{3.5}
     \\ 
     & 
     \textbf{\textit{A dog playing with a frisbee on the ground.}}
     &&&&&
     \\
     \cmidrule(lr){2-6}
     & 
     \textbf{1-B}: VinVL-base
     &
     \multirow{2}{*}{5}
     &
     \multirow{2}{*}{4}
     &
     \multirow{2}{*}{0.1}
     & 
     \multirow{2}{*}{4.4}
     \\ 
     &
     \textbf{\textit{A otter is laying on the sand next to two frisbees.}}
     &&&&&
     \\
     
     \cmidrule(lr){2-6}
     & 
     \textbf{1-C}: VinVL-large
     & 
     \multirow{2}{*}{4}
     &
     \multirow{2}{*}{3}
     &
     \multirow{2}{*}{0}
     & 
     \multirow{2}{*}{3.5}
     \\
     & 
     \textbf{\textit{A small animal laying on a rock with three frisbees.}}
     &&&&&
      \\ 
      
           \cmidrule(lr){1-6}
    \raisebox{0.0cm}{
     \multirow{6}{*}{
     {\includegraphics[width=0.21\textwidth]{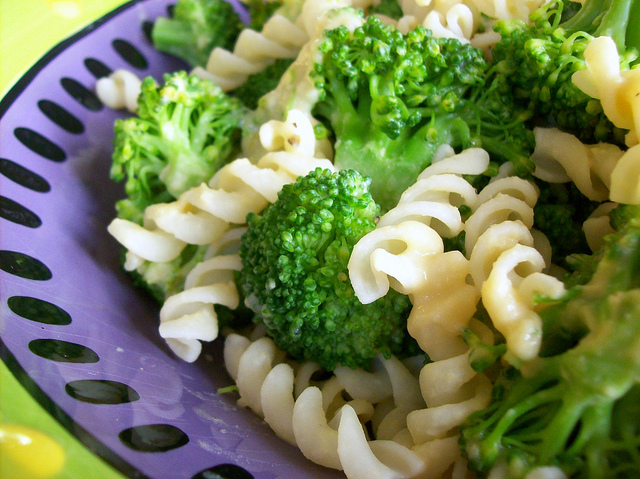}}
     }}
     & 
     \textbf{2-A}: Up-Down 
     &  
     \multirow{2}{*}{5}
     &
     \multirow{2}{*}{3}
     &
     \multirow{2}{*}{0}
     & 
     \multirow{2}{*}{4}
     \\ 
     & 
     \textbf{\textit{A close up of a plate of broccoli.}}
     &&&&&
     \\
     \cmidrule(lr){2-6}
     & 
     \textbf{2-B}: Unified-VLP, VinVL-base, VinVL-large
     &
     \multirow{2}{*}{4}
     &
     \multirow{2}{*}{4}
     &
     \multirow{2}{*}{0}
     & 
     \multirow{2}{*}{4}
     \\ 
     &
     \textbf{\textit{A plate of pasta and broccoli on a table.}}
     &&&&&
     \\
     
     \cmidrule(lr){2-6}
     & 
     \textbf{2-C}: Human
     & 
     \multirow{2}{*}{5}
     &
     \multirow{2}{*}{5}
     &
     \multirow{2}{*}{0.1}
     & 
     \multirow{2}{*}{4.9}
     \\
     & 
     \textbf{\textit{A multi colored dish with broccoli and white twisted pasta in it.}}
     &&&&&
     \\ 
      
           \cmidrule(lr){1-6}
    \raisebox{0.0cm}{
     \multirow{6}{*}{
     {\includegraphics[width=0.21\textwidth]{images/mscoco/COCO_val2014_000000565198.jpg}}
     }}
     & 
     \textbf{3-A}: Unified-VLP
     &  
     \multirow{2}{*}{3}
     &
     \multirow{2}{*}{4}
     &
     \multirow{2}{*}{0}
     & 
     \multirow{2}{*}{3.5}
     \\ 
     & 
     \textbf{\textit{A little girl holding a video game controller.}}
     &&&&&
     \\
     \cmidrule(lr){2-6}
     & 
     \textbf{3-B}: VinVL-large
     &
     \multirow{2}{*}{4}
     &
     \multirow{2}{*}{5}
     &
     \multirow{2}{*}{0}
     & 
     \multirow{2}{*}{4.5}
     \\ 
     &
     \textbf{\textit{A little girl is blow drying her hair on a couch.}}
     &&&&&
     \\
     
     \cmidrule(lr){2-6}
     & 
     \textbf{3-C}: Human
     & 
     \multirow{2}{*}{5}
     &
     \multirow{2}{*}{5}
     &
     \multirow{2}{*}{0}
     & 
     \multirow{2}{*}{5}
     \\
     & 
     \textbf{\textit{A little girl holding a blow dryer next to her head.}}
     &&&&&
     \\ 
           \cmidrule(lr){1-7}
    \raisebox{0.0cm}{
     \multirow{6}{*}{
     {\includegraphics[width=0.21\textwidth]{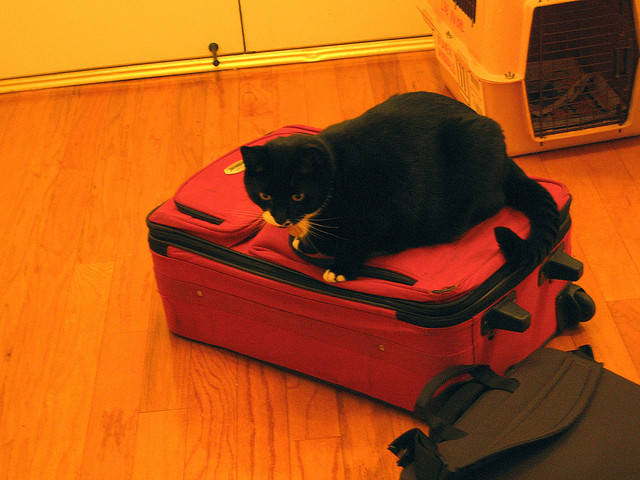}}
     }}
     & 
     \textbf{4-A}: Up-Down 
     &  
     \multirow{2}{*}{3}
     &
     \multirow{2}{*}{5}
     &
     \multirow{2}{*}{0}
     & 
     \multirow{2}{*}{4}
     \\ 
     & 
     \textbf{\textit{A black cat laying in a red suitcase.}}
     &&&&&
     \\
     \cmidrule(lr){2-6}
     & 
     \textbf{4-B}: Unified-VLP, VinVL-base, VinVL-large
     &
     \multirow{2}{*}{5}
     &
     \multirow{2}{*}{5}
     &
     \multirow{2}{*}{0}
     & 
     \multirow{2}{*}{5}
     \\ 
     &
     \textbf{\textit{A black cat sitting on top of a red suitcase.}}
     &&&&&
     \\
     
     \cmidrule(lr){2-7}
     & 
     \textbf{4-C}: Human
     & 
     \multirow{2}{*}{4}
     &
     \multirow{2}{*}{5}
     &
     \multirow{2}{*}{0}
     & 
     \multirow{2}{*}{4.5}
     \\
     & 
     \textbf{\textit{A large black cat laying on top of a pink piece of luggage.}}
     &&&&&
     \\ 
     
           \cmidrule(lr){1-6}
    \raisebox{0.0cm}{
     \multirow{6}{*}{
     {\includegraphics[width=0.21\textwidth]{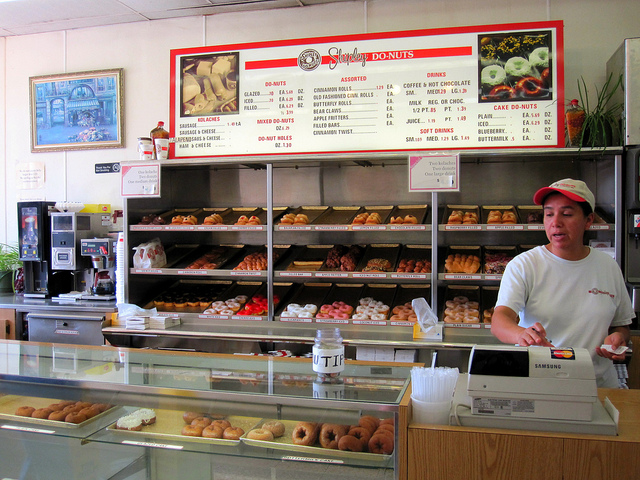}}
     }}
     & 
     \textbf{5-A}: Up-Down, Unified-VLP
     &  
     \multirow{2}{*}{3}
     &
     \multirow{2}{*}{2}
     &
     \multirow{2}{*}{0}
     & 
     \multirow{2}{*}{2.5}
     \\ 
     & 
     \textbf{\textit{A man standing in front of a display of donuts.}}
     &&&&&
     \\
     \cmidrule(lr){2-6}
     & 
     \textbf{5-B}: VinVL-large
     &
     \multirow{2}{*}{5}
     &
     \multirow{2}{*}{3}
     &
     \multirow{2}{*}{0}
     & 
     \multirow{2}{*}{4}
     \\ 
     &
     \textbf{\textit{A woman standing behind a counter at a donut shop.}}
     &&&&&
     \\
     
     \cmidrule(lr){2-6}
     & 
     \textbf{5-C}: Human
     & 
     \multirow{2}{*}{5}
     &
     \multirow{2}{*}{5}
     &
     \multirow{2}{*}{0.3}
     & 
     \multirow{2}{*}{4.7}
     \\
     & 
     \textbf{\textit{Woman selling doughnuts with doughnut stock in the background.}}
     &&&&&
     \\

      \bottomrule
      \end{tabular}
      \caption{Example evaluations of machine- and human-generated captions. None of these captions get penalties in conciseness and inclusive language. Evaluated captioning models are described in \S\ref{sec:evaluated-captions}. All MSCOCO images are provided under a Creative Commons Attribution 4.0 License \cite{mscoco}.}
      \vspace{-0.4cm}
      \label{table:eval_examples}
      \end{center}
\end{table*}

\subsubsection{Main Scores}
\label{sec:main_scores}
The two main scores are assessed in the scale of 1--5.
They balance information accuracy and coverage.
\paragraph{Precision}
Precision (P) measures how precise the caption is given the image.
For instance, Caption 1-B in Table \ref{table:eval_examples} is perfectly precise, while 1-A (\textit{dog} vs.\ \textit{otter}, \textit{one} vs.\ \textit{two frisbees}) and 1-C (\textit{three} vs.\ \textit{two frisbees}) are not precise.
Precision guards against hallucinations from the language model (\textit{table} in 2-B) that are known to be common failures of image captioning models \cite{rohrbach-etal-2018-object}.
The score of 4 is reserved for relatively minor issues, such as attributes that are almost correct (e.g., \textit{pink} vs.\ \textit{red} in 4-C, Table \ref{table:eval_examples}) or cases where the caption does not contradict  the image but is not guaranteed to be true (e.g., it is unclear whether the girl is sitting on a couch in 3-B).
In addition to objects themselves, precision deals with information like properties, attributes, occasions, locations, and relations between objects (e.g., \textit{in a red suitcase} vs.\ \textit{on a red suitcase} in 4-A).

\paragraph{Recall}
Recall (R) measures how much of the salient information (e.g., objects, attributes, and relations) from the image is covered by the caption.
This includes color (e.g., color of the frisbees in 1-A, 1-B, and 1-C) and guards against generic, uninformative captions that machines tend to produce \cite{wang2020spiceu}.
For instance, an otter is a \textit{small animal}, and thus \textit{small animal} is \textit{precise} (1-C); however, it is much less informative (and less natural; \citealp{entry-level-categories}) than saying an otter.
Similarly, Caption 5-B only says a woman is standing behind a counter at a donut shop, but she is selling donuts, not buying or looking at donuts, which is salient information from the picture.
We do not take a point off if missing information is already expected from the caption (e.g., a double-decker bus is typically red).
We often find it useful to take a generative approach when evaluating recall: \textit{what image does the caption lead us to imagine?} When the caption entails many potential images that substantially diverge from the given image, the recall score should be low.\footnote{Prior work found recall (or \textit{specificity}) can vary across cultures or languages \cite{MiltenburgEV17}. We focus on the English language in this work.}

\subsubsection{Penalties}
\label{sec:penalties}
\paragraph{Fluency}
Fluency (Flu.) measures the quality of captions as English text regardless of the given image.
Initially, we scored fluency in the scale of 1--5, similar to P and R, but we found most captions from modern neural network models were highly fluent.
Thus, we instead decided to take points off from the average of P and R if there's a fluency problem to account for minor issues that are much less problematic than losing one P/R point.
The four annotators had extensive discussions and developed rubrics for fluency.
Similar to recent work on professional evaluations for machine translation \cite{freitag2021experts}, we evaluated under the following principle: if a fluency problem is expected to be easily corrected by a text postprocessing algorithm (e.g., grammatical error correction: \citealp{yuan-briscoe-2016-grammatical,sakaguchi-etal-2017-grammatical}), 
the penalty should be 0.1. 
This includes obvious misspellings or grammatical errors (e.g., \textit{A otter} in 1-B) and missing determiners/hyphens (\textit{multi colored} in 2-C).
0.5+ points were subtracted for more severe problems, such as duplication (e.g., \textit{A display case of donuts and doughnuts}), ambiguity (e.g., \textit{A cat is on a table with a cloth on \textbf{it}}), and broken sentences (e.g., \textit{A large concrete sign small buildings behind it.}).
See Table \ref{tab:fluency_rubrics} in \S\ref{sec:fluencyrubrics} for more extensive fluency rubrics.
Note that the average fluency penalty was 0.01; this confirms that fluency is no longer crucial in ranking models for MSCOCO captioning and contrasts with human evaluations previously done for older captioning models.

\paragraph{Conciseness}
The scores so far do not take into account conciseness of captions. 
Specifically, a model could simply increase all scores by describing every detail in a picture.
For instance, the following caption is overly repetitive: \textit{a woman lying on her back with knees bent on a beach towel under a multicolored, striped beach umbrella, surrounded by sand, and with clear blue sky above}.
We subtract 0.5 points for these captions.
Note that most machine captions were short, and this penalty was only applied to two human-generated captions.
It might become more crucial for future models with a more powerful object detection module that catches many objects in the picture.

\paragraph{Inclusive Language}
We found that some instances substantially diverge from inclusive language when humans are described \cite{van-miltenburg-2020-image}, raising a concern for downstream applications.
In these cases, we added a penalty: 0.5 points were deducted for a subjective comment about appearance (e.g., \textit{very pretty girl}), and 2 points for more severe problems (e.g., \textit{beautiful breasts}).

\subsubsection{Rules of \thumb}
In our development phase, we established the following additional rules to clarify our annotation scheme.
\paragraph{Avoiding Double Penalties}
When an error is accounted for in precision, we correct the error before scoring the recall, thereby avoiding penalizing the precision and recall for the same mistake.
For example, P=3 is given to Caption 1-A in Table \ref{table:eval_examples} because of its wrong detection (\textit{dog} vs.\ \textit{otter}; \textit{one} vs.\ \textit{two frisbees}), but we score the recall assuming that the caption is now \textit{an \textbf{otter} playing with two frisbees on the ground}.
This ensures that a generic, useless caption, such as \textit{there is something on something} (P=5, R=1), would be ranked considerably lower than \textit{a dog on the beach with two pink and yellow frisbees} (P=3, R=5).
Similarly, the wrong detection in 5-A (\textit{man} vs.\ \textit{woman}) is handled only in precision.
Note that such error correction is not applicable to hallucinations because there is no alignment between a part of the image and a hallucinated object (e.g., \textit{table} in 2-B).
This rule departs from the definition of recall in SPICE \cite{SPICE16}, an automatic metric that measures the $F_1$ score in scene graphs predicted from reference and generated captions; their alignment is limited to WordNet synonyms \cite{miller1995wordnet}.
This means that classifying an otter as a dog or even a small animal would result in cascading errors both in precision and recall, overrating captions that completely overlook the otter or ones that make a more severe classification error (e.g., miscategorize the otter as a car, compared to a dog).

\paragraph{Object Counts as Attributes}
All counts are considered as object attributes, and wrong counts are handled in precision.
This simplifies the distinction between precision and recall.
For instance, both \textit{a frisbee} (1-A) and \textit{three frisbees} (1-C) are precision problems, while saying \textit{some frisbees} would be a recall problem when it is clear that there are exactly two frisbees.
Note that this is in line with SPICE, which treats object counts as attributes in a scene graph, rather than duplicating a scene graph for every instance of an object \cite{SPICE16}.

\paragraph{Black and White Photo}
MSCOCO contains black and white or gray-scale pictures.
Some captions explicitly mention that they are black and white, but we disregard this difference in our evaluations.
The crowdsource instructions for creating reference captions do not specify such cases \cite{mscoco-captioning}. Further, we can potentially run postprocessing to determine whether it is black and white to modify the caption accordingly, depending on the downstream usage.

\paragraph{Text Processing}
Image captioning models often differ slightly in text preprocessing.
As a result, we found that generated captions were sometimes slightly different in format (e.g., tokenized or detokenized; lowercased or not).
For better reproducibility, we follow the spirit of \textsc{SacreBLEU} \cite{post-2018-call}, which has become the standard package to compute BLEU scores for machine translation: all evaluations, including automatic metrics, should be done on clean, untokenized text, independently of preprocessing design choices.  
We apply the following minimal postprocessing to the model outputs and human captions.
\begin{itemize}[noitemsep, leftmargin=*, topsep=0pt]
\item Remove unnecessary spaces at the start or end of every caption. 
\item Uppercase the first letter.
\item Add a period at the end if it doesn't exist, and remove a space before a period if any.
\end{itemize}
We keep the postprocessing minimal for this work and encourage future model developers to follow the standard practice in machine translation: every model has to output clean, truecased, untokenized text that is ready to be used in downstream modules.
This also improves the transparency and reproducibility of automated evaluations \cite{post-2018-call}.

\subsection{Evaluated Captions}
\label{sec:evaluated-captions}
We evaluated the following four strong models from the literature as well as human-generated captions.
They share similar pipeline structure: object detection followed by crossmodal caption generation.
They vary in model architecture, (pre)training data, model size, and (pre)training objective. Evaluating captions from them will enable us to better understand what has been improved and what is still left to future captioning models.
\begin{itemize}[noitemsep, leftmargin=*, topsep=0pt]
\item \textbf{Up-Down} \cite{Anderson2017up-down} trains Faster R-CNN \cite{fasterRCNN} on the Visual Genome datset \cite{krishnavisualgenome} for object detection. It then uses an LSTM-based crossmodal generation model. 
\item \textbf{Unified-VLP} \cite{Unified-VLP} uses the same object detection model as Up-Down. The transformer-based generation model is initialized with base-sized BERT \cite{devlins2019bert} and further pretrained with 3M images from Conceptual Captions \cite{sharma-etal-2018-conceptual}.
\item \textbf{VinVL-base} and \textbf{VinVL-large} \cite{zhang2021vinvl} train a larger-scale object detection model with the ResNeXt-152 C4 architecture \cite{C4} on ImageNet \cite{deng2009imagenet}. The transformer generation model is initialized with BERT and pretrained with 5.7M images.
\item \textbf{Human} randomly selects one from the five human-generated reference captions in MSCOCO. Those captions were created by crowdworkers on Amazon Mechanical Turk \cite{mscoco-captioning}.
\end{itemize}
Further details are described in \S\ref{app:evaluated-captions} of Appendix.
\begin{table*}[t]
\small
\centering
\addtolength{\tabcolsep}{-2.55pt}  
\def\arraystretch{1.2}
\begin{tabular}{@{} lcccccc @{\hspace{0.25cm}} | c@{\hspace{0.15cm}}c@{\hspace{0.15cm}}c@{\hspace{0.15cm}}c@{\hspace{0.15cm}}c@{\hspace{0.15cm}}c@{\hspace{0.15cm}}c @{}}
\toprule
 & \multicolumn{6}{c|}{\textbf{\thumb 1.0}} &   \multicolumn{7}{c}{\textbf{Automatic Metrics}} \\
\textbf{Model}& P$\uparrow$ & R$\uparrow$ & Flu.$\downarrow$ & Con.$\downarrow$ & Inc.$\downarrow$ & Total$\uparrow$ & BLEU & ROUGE & BERT-S & SPICE & CIDEr & CLIP-S & RefCLIP-S\\
\hline
Human & $4.82$ & $4.35$ & $0.019$ & $0.02$ & $0.00$ & $\mathbf{4.56}_{-0.03}^{+0.03}$ &  
$26.2$ & $50.4$ & $0.938$ & $23.7$ & $111.5$ & $0.791$ & $0.834$
\\
VinVL-large & $4.54$ & $3.97$ & $0.005$  & $0.00$ & $0.00$ & $4.25_{-0.04}^{+0.04}$ &
$33.3$& $56.5$ & $0.946$ & $26.4$ & $141.8$ & $0.784$ & $0.834$
\\
VinVL-base & $4.47$ & $3.95$ & $0.001$ & $0.00$ & $0.00$ &  $4.21_{-0.04}^{+0.04}$ & 
$32.3$ & $55.9$ & $0.945$ & $25.6$ & $138.4$ & $0.779$ & $0.830$
\\
Unified-VLP & $4.35$ & $3.77$ & $0.004$ & $0.00$ & $0.00$ & $4.06_{-0.04}^{+0.04}$ & 
$31.6$ & $55.8$ & $0.945$ & $24.3$ & $128.5$ & $0.771$ & $0.821$
\\
Up-Down & $4.29$ & $3.50$ & $0.014$ & $0.00$ & $0.00$ & $3.88_{-0.05}^{+0.05}$ &
$28.4$& $52.2$ & $0.939$ & $21.0$ & $110.7$ & $0.746$ & $0.803$
\\
\bottomrule
\end{tabular}
\caption{Performance of image captioning models with respect to \thumb 1.0 (left) and automatic metrics (right). All scores are averaged over 500 images randomly sampled from the Karpathy test split. P: precision; R: recall; Flu.: fluency; Con.: conciseness; Inc.: inclusive language. 90\% confidence intervals for total scores are calculated by bootstrapping \cite{koehn-2004-statistical}. All reference-based metrics take as input the same four crowdsourced captions that are not used in Human for fair comparisons.
\textbf{\thumb 1.0 scores Human substantially higher than the machines, unlike all automatic metrics.}
}
\label{tab:model_results}
\end{table*}

\section{Results and Analysis}
We present results and analysis from our evaluations.
Our transparent evaluations facilitate assessments and analysis of both captioning models (\S\ref{sec:model_comparisons}) and automatic metrics (\S\ref{sec:metric_comparisons}).
\subsection{Comparing Models}
\label{sec:model_comparisons}
Seen in Table \ref{tab:model_results} (left section) is the model performance that is averaged over the 500 test images and broken down by the rubric categories.
Overall, Human substantially outperforms all machines in the P, R, and total scores.
In particular, we see a large gap between Human and the machines in recall (e.g., Human 4.35 vs.\ VinVL-large 3.97). 
This contrasts with the automatic metric-based ranking of the MSCOCO leaderboard, where Human is ranked at the 250th place.\footnote{The official leaderboard ranks submissions using CIDEr \cite{cider2015} with 40 references on the hidden test data. We use the public Karpathy test split instead, but we suspect the same pattern would hold on the hidden data as well, given the large gap between machines and Human.}
This result questions claims about human parity or superhuman performance on MSCOCO image captioning.
The four machine captioning models are ranked in the expected order, though the small difference between VinVL-large and VinVL-base suggests that simply scaling up models would not lead to a substantial improvement.
We see that the three models that are initialized with pretrained BERT (VinVL-large/base, Unified-VLP) are particularly fluent, but the problem is small in the other models as well.

While we compute representative, total scores, our transparent rubrics allow for adjusting weighting of the categories depending on the application of interest.
For instance, in the social media domain, recall can be more important than precision to make captions engaging to users \cite{Shuster2019}.
To assess the models independently of these aggregation decisions, we count the number of times when each model outperforms/underperforms all the others both in P and R (\textit{strictly} best/worst, Table \ref{tab:strict_numbers}). 
We see patterns consistent with Table \ref{tab:model_results}.
For example, Human is most likely to be strictly best and least likely to be strictly worst.
This suggests that machine captioning models would still fall short of crowdworkers in a wide range of downstream scenarios.

\begin{table}[h]
\small
\centering
\addtolength{\tabcolsep}{-3.0pt}  
\def\arraystretch{1.2}
\begin{tabular}{@{} lccccc @{}}
\toprule
\textbf{Model}& Human &  Vin-large & Vin-base & U-VLP & Up-Down \\
\hline
\# Best $\uparrow$ & \textbf{327}  & 180 & 161 & 112 & 74   \\
\# Worst $\downarrow$ & \textbf{65} & 128  & 150 & 190 & 269 \\
\bottomrule
\end{tabular}
\caption{\# times when each captioning model is \textit{strictly} best/worst in the caption set (i.e., best/worst both in precision and recall).}
\label{tab:strict_numbers}
\end{table}

\subsection{Comparing Automatic Metrics}
\label{sec:metric_comparisons}
While carefully-designed human judgments like ours should be considered more reliable, automatic metrics allow for faster development cycles.
Our transparent evaluations can also be used to analyze how these automatic metrics correlate with different aspects of image captioning.
Table \ref{tab:model_results} (right section) shows automatic scores of the captioning models over 7 popular metrics for image captioning. 
CLIP-S \cite{clipscore} is a \textit{referenceless} metric that uses image features from CLIP \cite{clip2021}, a crossmodal retrieval model trained on 400M image-caption pairs from the web.
RefCLIP-S augments CLIP-S with similarities between the generated and reference captions.
All other metrics, such as SPICE \cite{SPICE16} and CIDEr \cite{cider2015}, only use reference captions without image features.

These automatic metrics generally agree with our evaluations in ranking the four machines, but completely disagree in the assessment of Human.
Most metrics rank Human near the bottom, showing that they are not reliable in evaluating high-quality, human-generated captions.
The two metrics with powerful image and text features (CLIP-S and RefCLIP-S) give high scores to Human compared to the other metrics, but they still fail to score Human substantially higher than VinVL-large.
This suggests that automatic metrics should be regularly updated as our models become stronger (and perhaps more similar to humans), and raises a significant concern about the current practice that fixes evaluation metrics over time \cite{billboard}.
\begin{table}[h]
\small
\centering
\addtolength{\tabcolsep}{-2.6pt}  
\def\arraystretch{1.2}
\begin{tabular}{@{} lccc|ccccc @{}}
\toprule
&    \multicolumn{3}{c|}{w/o Human} & \multicolumn{3}{c}{w/ Human}  \\
\textbf{Metric} & P  & R & Total & P & R & Total \\
\hline
RefCLIP-S & $0.34$ & $\mathbf{0.27}$ & $\mathbf{0.44}$ & $0.31$ & $0.26$ &$\mathbf{0.41}^{+0.05}_{-0.05}$ \\
RefOnlyC & $\textbf{0.42}$ & $0.14$ &  $0.41$ & $\mathbf{0.37}$ & $0.11$ & $0.34^{+0.04}_{-0.05}$\\
CLIP-S & $0.18$ & $\mathbf{0.27}$ &  $0.32$ & $0.17$ & $\mathbf{0.28}$ & $0.32^{+0.05}_{-0.05}$\\
CIDEr &  $0.27$ & $0.18$ & $0.33$ & $0.21$ & $0.11$ & $0.23^{+0.04}_{-0.04}$ \\
BERT-S &  $0.27$ & $0.18$  &  $0.33$ & $0.20$ & $0.10$ & $0.21^{+0.04}_{-0.04}$ \\
SPICE &  $0.26$ & $0.15$ & $0.30$ & $0.20$ & $0.09$ & $0.21^{+0.04}_{-0.04}$\\
ROUGE-L & $0.26$ & $0.17$ & $0.31$ & $0.18$ & $0.07$ & $0.18^{+0.04}_{-0.04}$ \\
BLEU  & $0.21$ & $0.13$ & $0.25$ & $0.15$ & $0.04$ & $0.13^{+0.04}_{-0.04}$\\
\bottomrule
\end{tabular}
\caption{Instance-level correlations of automatic evaluation scores. RefCLIP-S and CLIP-S use image features unlike the others, and all but CLIP-S require references. All of these reference-based metrics use the same subset of four captions as in Table \ref{tab:model_results} that exclude Human. All metrics had correlations lower than 0.1 for fluency.}
\label{tab:metric_results}
\end{table}

Seen in Table \ref{tab:metric_results} are instance-level Pearson correlation scores between automatic scores and our evaluations.\footnote{Instance-level Pearson correlations with human judgments were often computed in prior work to compare automatic metrics for image captioning (e.g., \citealp{clipscore}). An alternative is system-level correlations, but they would be uninformative with five systems only.}
We also add an ablation study: RefOnlyC removes image features from RefCLIP-S to quantify the effect of image features.
We consider two types of scenarios: one \textit{with} Human and one \textit{without}.
Correlations drop from the latter to the former for all metrics and aspects except CLIP-S, again showing that the metrics are not reliable in assessing human-generated captions.
Interestingly, CLIP-S correlates best in recall (0.28 w/ Human) but suffers in precision (0.17 w/ Human).
RefOnlyC, in contrast, achieves the best correlations in P at the expense of R.
RefCLIP-S balances the two and achieves the best correlation in total scores.
This indicates that the CLIP image features particularly help assess coverage of salient information that can be ignored in some reference captions from crowdworkers.\footnote{The low recall correlations of reference-only metrics can be partly because the maximum (as opposed to minimum or average) is typically taken over multiple reference captions (e.g., BERTScore, \citealp{bertscore}). 
Nevertheless, this alone does not explain the recall gap from image-based metrics because RefCLIP-S also takes the maximum score over all references.
Future work can explore the relation between precision/recall and different treatments of multiple references.}
Prior work \cite{clipscore} found that SPICE can still improve correlations when combined with CLIP-S, even though CLIP-S better correlates with human judgments than SPICE.
This implies that image-based and reference-only metrics capture different aspects of image captioning.
Our analysis indeed agrees with their finding and, further, identifies that recall is one such aspect.
For an extensive description of these metrics and their configurations, see \S\ref{sec:decription_metrics}.

\subsection{Score Distributions}
\label{sec:score_distributions}
\begin{figure}[h]
\centering
    \includegraphics[scale=0.44]{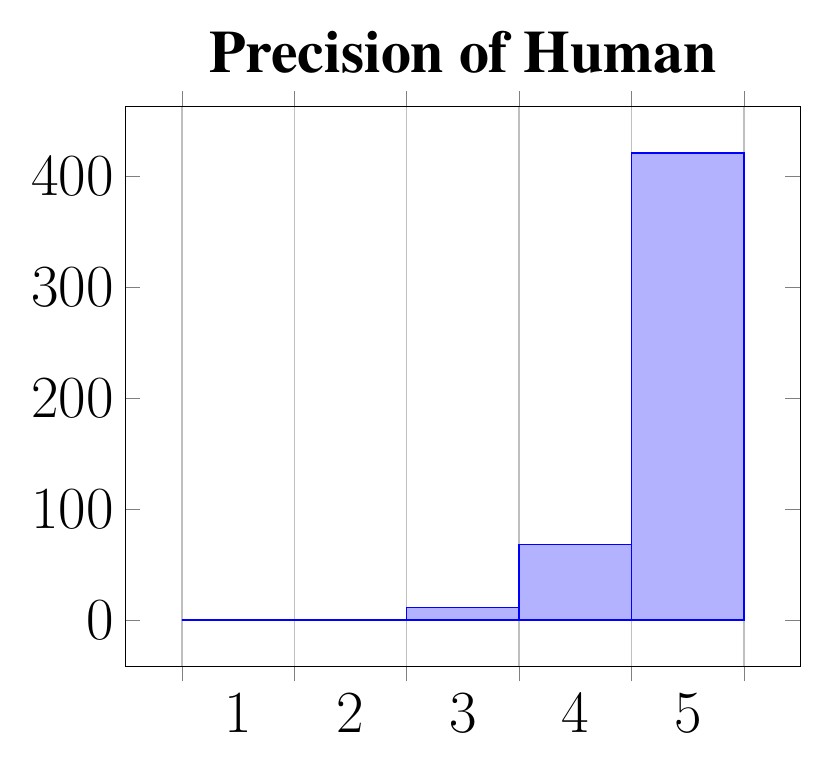}
    \includegraphics[scale=0.44]{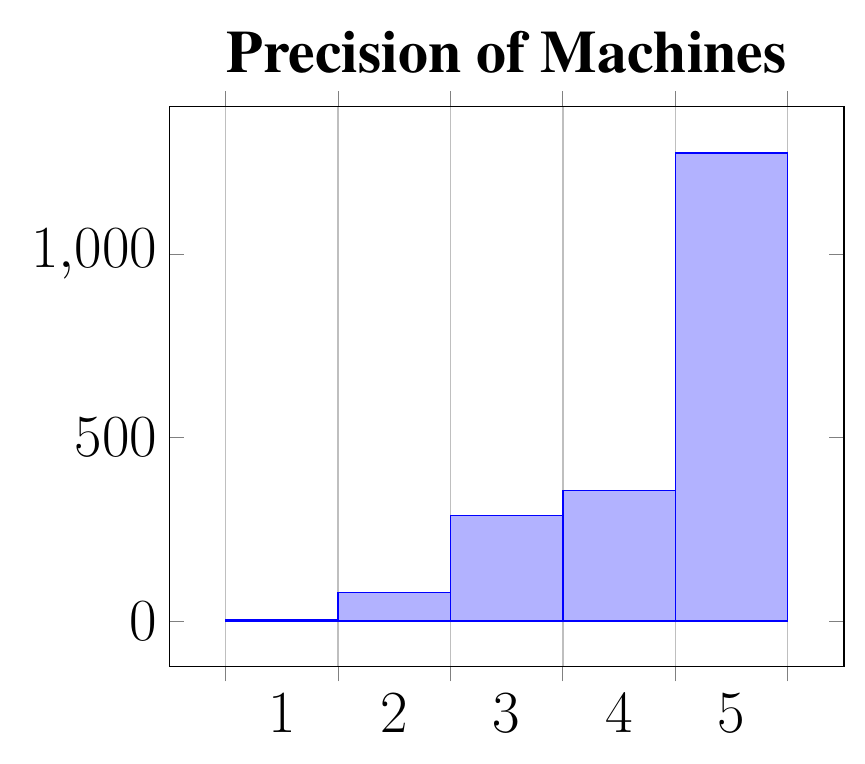}
    \includegraphics[scale=0.44]{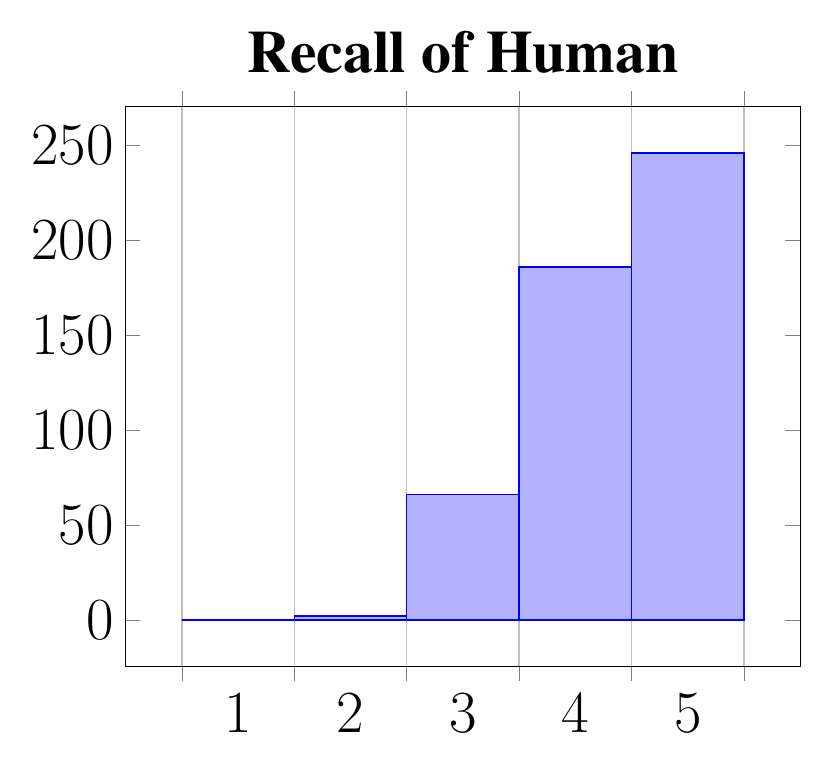}
    \includegraphics[scale=0.44]{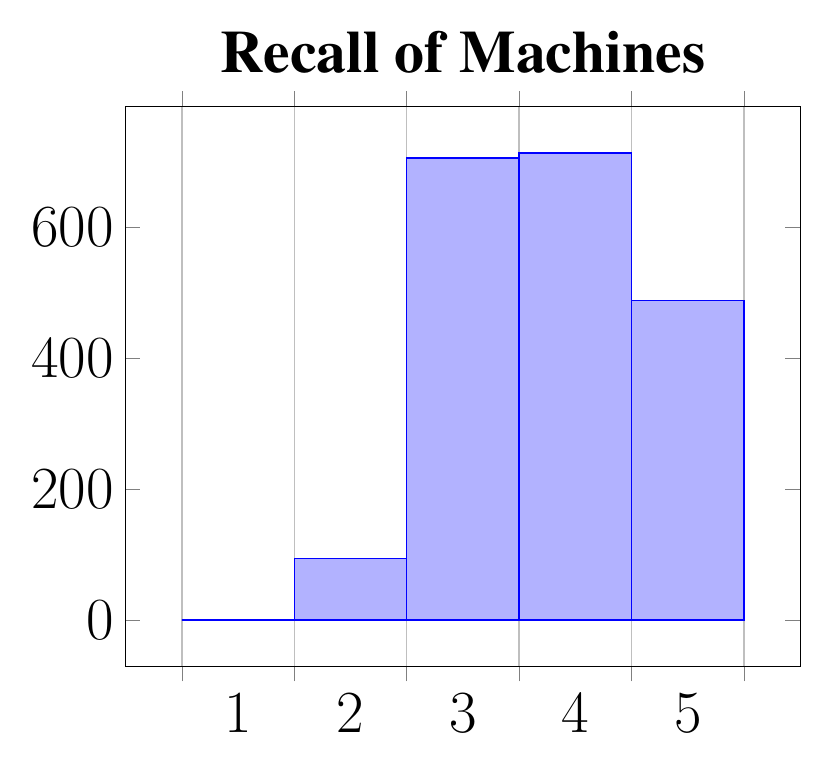}\\
    
\caption{Precision/recall histograms for human- and machine-generated captions.}
\label{fig:scores_hist}
\end{figure}

Seen in Fig.\ \ref{fig:scores_hist} are distributions of precision and recall scores for human and machine-generated captions.
We see that the precision distribution looks similar between Human and machines, but not recall. 
This provides further support for our claim that current machines fall short of humans particularly in recall. 

\subsection{Machine vs.\ Human Examples}
\begin{table*}[h!]
    \small
     \begin{center}
     \begin{tabular}{ @{}  c p{8.5cm}  c c c c  @{} }
     \toprule
      Image & Caption & P & R & Flu.& Total\\ 
    \cmidrule(lr){1-1}\cmidrule(lr){2-2}
    \cmidrule(lr){3-5}
    \cmidrule(lr){6-6}
    \raisebox{0.0cm}{
     \multirow{6}{*}{
     {\includegraphics[width=0.10\textwidth]{images/mscoco/COCO_val2014_000000361147.jpg}}
     }}
     & 
     \textbf{6-A}: Up-Down 
     &  
     \multirow{2}{*}{5}
     &
     \multirow{2}{*}{3}
     &
     \multirow{2}{*}{0}
     & 
     \multirow{2}{*}{4}
     \\ 
     & 
     \textbf{\textit{A man holding a tennis racquet on a tennis court.}}
     &&&&
     \\
     \cmidrule(lr){2-6}
     & 
     \textbf{6-B}: Unified-VLP, VinVL-base, VinVL-large
     &
     \multirow{2}{*}{5}
     &
     \multirow{2}{*}{3}
     &
     \multirow{2}{*}{0}
     & 
     \multirow{2}{*}{4}
     \\ 
     &
     \textbf{\textit{A man holding a tennis racket on a tennis court.}}
     &&&&
     \\
     
     \cmidrule(lr){2-6}
     & 
     \textbf{6-C}: Human
     & 
     \multirow{2}{*}{5}
     &
     \multirow{2}{*}{5}
     &
     \multirow{2}{*}{0}
     & 
     \multirow{2}{*}{5}
     \\ 
     &
     \textbf{\textit{A tennis player shows controlled excitement while a crowd watches.}}
     &&&&
     \\

           \cmidrule(lr){1-6}
     \multirow{10}{*}{\raisebox{-0.3cm}{
     {\includegraphics[width=0.21\textwidth]{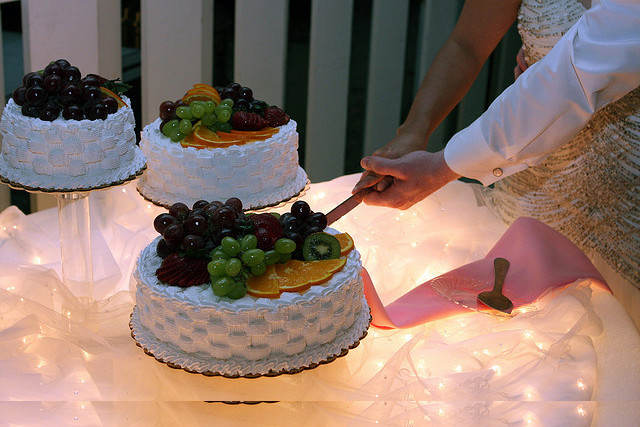}}
     }}
     & 
     \textbf{7-A}: Up-Down 
     &  
     \multirow{2}{*}{3}
     &
     \multirow{2}{*}{3}
     &
     \multirow{2}{*}{0}
     & 
     \multirow{2}{*}{3}
     \\ 
     & 
     \textbf{\textit{A person cutting a cake with a knife.}}
     &&
     \\
     \cmidrule(lr){2-6}
     & 
     \textbf{7-B}: Unified-VLP
     &
     \multirow{2}{*}{3}
     &
     \multirow{2}{*}{5}
     &
     \multirow{2}{*}{0}
     & 
     \multirow{2}{*}{4}
     \\ 
     &
     \textbf{\textit{A person cutting a wedding cake with a knife.}}
     &&&&
     \\
     \cmidrule(lr){2-6}
     & 
     \textbf{7-C}: VinVL-base
     &
     \multirow{2}{*}{5}
     &
     \multirow{2}{*}{3}
     &
     \multirow{2}{*}{0}
     & 
     \multirow{2}{*}{4}
     \\ 
     &
     \textbf{\textit{A couple of cakes on a table with a knife.}}
     &&&&
     \\
     \cmidrule(lr){2-6}
     & 
     \textbf{7-D}: VinVL-large
     & 
     \multirow{2}{*}{3}
     &
     \multirow{2}{*}{3}
     &
     \multirow{2}{*}{0}
     & 
     \multirow{2}{*}{3}
     \\
     & 
     \textbf{\textit{A woman cutting a cake with a knife.}}
     &&&&
     \\ 
     
     \cmidrule(lr){2-6}
     & 
     \textbf{7-E}: Human
     & 
     \multirow{2}{*}{5}
     &
     \multirow{2}{*}{5}
     &
     \multirow{2}{*}{0.1}
     & 
     \multirow{2}{*}{4.9}
     \\
     & 
     \textbf{\textit{Bride and grooms arms cutting the wedding cake with fruit on top.}}
     &&&&
     \\ 
      
           \cmidrule(lr){1-6}
    \raisebox{0.0cm}{
     \multirow{10}{*}{
     {\includegraphics[width=0.21\textwidth]{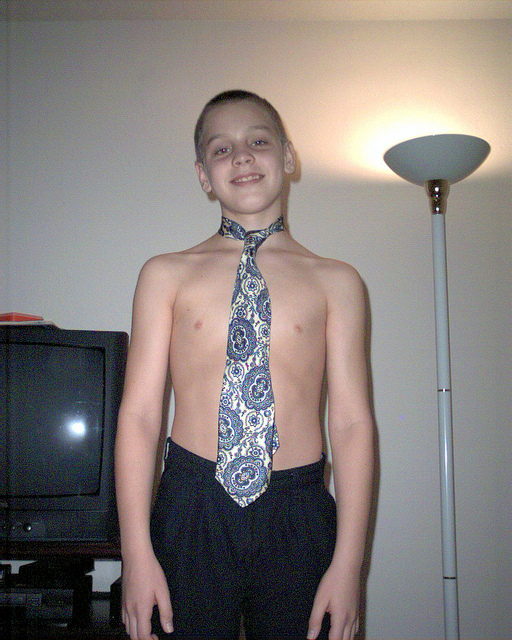}}
     }}
     & 
     \textbf{8-A}: Up-Down
     &  
     \multirow{2}{*}{3}
     &
     \multirow{2}{*}{3}
     &
     \multirow{2}{*}{0}
     & 
     \multirow{2}{*}{3}
     \\ 
     & 
     \textbf{\textit{A young boy wearing a blue shirt and a blue tie.}}
     &&&&
     \\
     \cmidrule(lr){2-6}
     & 
     \textbf{8-B}: Unified-VLP
     &
     \multirow{2}{*}{3}
     &
     \multirow{2}{*}{3}
     &
     \multirow{2}{*}{0}
     & 
     \multirow{2}{*}{3}
     \\ 
     &
     \textbf{\textit{A young boy wearing a shirt and a tie.}}
     &&&&
     \\
     
     \cmidrule(lr){2-6}
     & 
     \textbf{8-C}: VinVL-base
     & 
     \multirow{2}{*}{5}
     &
     \multirow{2}{*}{3}
     &
     \multirow{2}{*}{0}
     & 
     \multirow{2}{*}{4}
     \\
     & 
     \textbf{\textit{A young boy wearing a tie standing in front of a lamp.}}
     &&&&
     \\ 
     
     \cmidrule(lr){2-6}
     & 
     \textbf{8-D}: VinVL-large
     & 
     \multirow{2}{*}{3}
     &
     \multirow{2}{*}{3}
     &
     \multirow{2}{*}{0}
     & 
     \multirow{2}{*}{3}
     \\
     & 
     \textbf{\textit{A young man wearing a tie and a shirt.}}
     &&&&
     \\ 
     
     \cmidrule(lr){2-6}
     & 
     \textbf{8-E}: Human
     & 
     \multirow{2}{*}{4}
     &
     \multirow{2}{*}{5}
     &
     \multirow{2}{*}{0}
     & 
     \multirow{2}{*}{4.5}
     \\
     & 
     \textbf{\textit{A man wearing only a tie standing next to a lamp.}}
     &&&&
     \\ 
     
           \cmidrule(lr){1-6}
    \raisebox{0.0cm}{
     \multirow{10}{*}{
     {\includegraphics[width=0.20\textwidth]{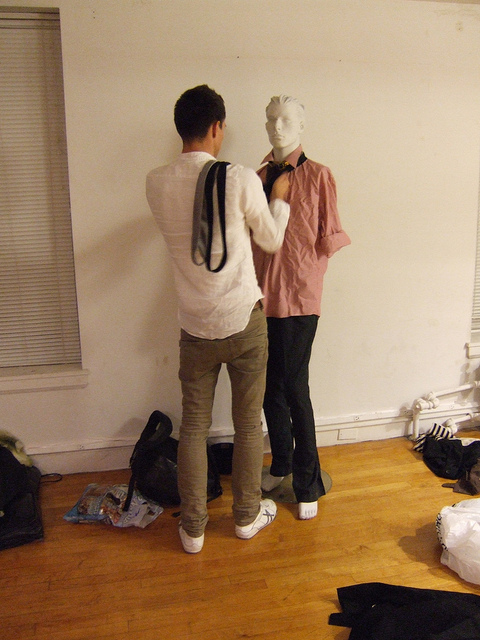}}
     }}
     & 
     \textbf{9-A}: Up-Down 
     &  
     \multirow{2}{*}{2}
     &
     \multirow{2}{*}{2}
     &
     \multirow{2}{*}{0}
     & 
     \multirow{2}{*}{2}
     \\ 
     & 
     \textbf{\textit{A couple of men standing next to each other.}}
     &&&&
     \\
     \cmidrule(lr){2-6}
     & 
     \textbf{9-B}: Unified-VL
     &
     \multirow{2}{*}{2}
     &
     \multirow{2}{*}{2}
     &
     \multirow{2}{*}{0}
     & 
     \multirow{2}{*}{2}
     \\ 
     &
     \textbf{\textit{Two men standing in a room.}}
     &&&&
     \\
     \cmidrule(lr){2-6}
     & 
     \textbf{9-C}: VinVL-base
     &
     \multirow{2}{*}{2}
     &
     \multirow{2}{*}{2}
     &
     \multirow{2}{*}{0}
     & 
     \multirow{2}{*}{2}
     \\ 
     &
     \textbf{\textit{A couple of men standing in a room.}}
     &&&&
     \\
     \cmidrule(lr){2-6}
     & 
     \textbf{9-D}: VinVL-large
     &
     \multirow{2}{*}{2}
     &
     \multirow{2}{*}{2}
     &
     \multirow{2}{*}{0}
     & 
     \multirow{2}{*}{2}
     \\ 
     &
     \textbf{\textit{Two men standing next to each other in a room.}}
     &&&&
     \\
     \cmidrule(lr){2-6}
     & 
     \textbf{9-E}: Human
     &
     \multirow{2}{*}{5}
     &
     \multirow{2}{*}{3}
     &
     \multirow{2}{*}{0}
     & 
     \multirow{2}{*}{4}
     \\ 
     &
     \textbf{\textit{A man standing next to a dummy wearing clothes.}}
     &&&&
     \\

      \bottomrule
      \end{tabular}
      \caption{Examples that contrast machine- and human-generated captions. All machine-generated captions overlook or misinterpret salient information: the excitement the tennis player expresses, the bride and groom cutting a wedding cake, the boy not wearing a shirt, and the man putting a tie on a dummy.
      None of these captions are penalized for conciseness or inclusive language.
      See \S\ref{sec:additional_machine_vs_human} in Appendix for more examples.}
      \label{table:machine_vs_human}
      \end{center}
\end{table*}

Table \ref{table:machine_vs_human} provides examples that contrast machine- and human-generated captions.
We see that machine-generated captions ignore salient information or make critical errors for these images.
These problems often occur in relatively rare cases: a tennis player is showing excitement rather than hitting a ball; a bride and groom are cutting a wedding cake; a boy is wearing a tie without a shirt; a man is putting clothing and a tie on a dummy instead of a person.
But these situations are exactly the most important information because of their \textit{atypicality} \cite{feinglass-yang-2021-smurf}.
This illustrates fundamental problems of current image captioning models that are left to future work.

\section{Related Work}
\paragraph{Human Evaluations for Image Captioning}
Several prior works conducted human evaluations for image captioning with varying models, datasets, and annotation schemes.
Much work used crowdworkers from Amazon Mechanical Turk on Flickr-based datasets, including the PASCAL \cite{rashtchian-etal-2010-collecting}, Flickr8k/30k \cite{flicker8k-expert-crowd, flicker30k}, and MSCOCO datasets.
Annotators scored the overall quality directly \cite{babytalk2011, flicker8k-expert-crowd}, pairwise \cite{cider2015}, or along multiple dimensions, such as truthfulness/correctness \cite{yatskar-etal-2014-see, SPICE16}, thoroughness \cite{images-to-sentences15}, relevance \cite{yang-etal-2011-corpus, li-etal-2011-composing}, and grammaticality/readability \cite{mitchell-etal-2012-midge, elliott-keller-2013-image}.
 There are similarities between our rubrics and previous annotations, but our framework defines every dimension in a decomposable way through discussions among all annotators, while focusing on outputs from strong models currently available.
Apart from these conventional Flickr-based datasets, some other work evaluated image captions for social media (engagingness, \citealp{Shuster2019}; accessibility for Twitter users with vision impairments, \citealp{Gleason19, GleasonCHI20}) and news articles \cite{Biten19}.
Our transparent evaluations would enable us to adjust the aggregation method based on the nature of downstream applications.
More specializing categories can be added for these applications in later versions (e.g., \thumb 2.0).

\paragraph{Human Evaluations for Other Generation Tasks}
Much previous work explored human evaluations for other language generation tasks than image captioning.
The WMT shared task \cite{akhbardeh-etal-2021-findings} conducts human evaluations of state-of-the-art machine translation systems every year; participants or crowdworkers directly rate a translation in a 100-point scale, which is a method developed by \citet{graham-etal-2013-continuous, graham-etal-2014-machine, GrahamBMZ17}.
\genie takes a similar approach but hosts human evaluations in leaderboards for machine translation, summarization, and commonsense reasoning \cite{genie}.
\citet{kryscinski-etal-2019-neural} and \citet{fabbri2021summeval} assessed many summarization models in a similar annotation scheme to the DUC 2006/2007 evaluations \cite{Dang2006OverviewOD}. 
Our transparent evaluation framework is inspired by rubric-based machine translation judgments by professional translators \cite{freitag2021experts}, which resulted in different system rankings than the WMT evaluations.
As top-performing models and automatic metrics are becoming increasingly similar across various natural language generation tasks, our findings on image captioning may be useful for other generation tasks as well.

\section{Conclusion}
We developed \thumb 1.0, transparent evaluations for the MSCOCO image captioning task.
We refined our rubrics through extensive discussions among all annotators, and ensured the high quality by two-stage annotations.
Our evaluations demonstrated critical limitations of current image captioning models and automatic metrics.
While recent image-based metrics show promising improvements, they are still unreliable in assessing high-quality captions from crowdworkers.
We hope that our annotation data will help future development of better captioning models and automatic metrics, and this work will become a basis for transparent human evaluations for the image captioning task and beyond.

\section*{Acknowledgments}
We thank Jack Hessel, Daniel Khashabi, Rowan Zellers, Noriyuki Kojima, Desmond Elliott, and Koji Shiono as well as the anonymous reviewers for their helpful feedback on this work.
We also thank Jack Hessel, Xiujun Li, Ani Kembhavi, Eric Kolve, and Luowei Zhou for their help with captioning models and automatic evaluation metrics. This work was supported in part by the DARPA MCS program through NIWC Pacific (N66001-19-2-4031) and  Google Cloud Compute.

\bibliography{custom}

\begin{thebibliography}{69}
\expandafter\ifx\csname natexlab\endcsname\relax\def\natexlab#1{#1}\fi

\bibitem[{Aditya et~al.(2015)Aditya, Yang, Baral, Ferm{\"{u}}ller, and
  Aloimonos}]{images-to-sentences15}
Somak Aditya, Yezhou Yang, Chitta Baral, Cornelia Ferm{\"{u}}ller, and Yiannis
  Aloimonos. 2015.
\newblock \href {http://arxiv.org/abs/1511.03292} {From images to sentences
  through scene description graphs using commonsense reasoning and knowledge}.

\bibitem[{Akhbardeh et~al.(2021)Akhbardeh, Arkhangorodsky, Biesialska, Bojar,
  Chatterjee, Chaudhary, Costa-jussa, Espa{\~n}a-Bonet, Fan, Federmann,
  Freitag, Graham, Grundkiewicz, Haddow, Harter, Heafield, Homan, Huck,
  Amponsah-Kaakyire, Kasai, Khashabi, Knight, Kocmi, Koehn, Lourie, Monz,
  Morishita, Nagata, Nagesh, Nakazawa, Negri, Pal, Tapo, Turchi, Vydrin, and
  Zampieri}]{akhbardeh-etal-2021-findings}
Farhad Akhbardeh, Arkady Arkhangorodsky, Magdalena Biesialska, Ond{\v{r}}ej
  Bojar, Rajen Chatterjee, Vishrav Chaudhary, Marta~R. Costa-jussa, Cristina
  Espa{\~n}a-Bonet, Angela Fan, Christian Federmann, Markus Freitag, Yvette
  Graham, Roman Grundkiewicz, Barry Haddow, Leonie Harter, Kenneth Heafield,
  Christopher Homan, Matthias Huck, Kwabena Amponsah-Kaakyire, Jungo Kasai,
  Daniel Khashabi, Kevin Knight, Tom Kocmi, Philipp Koehn, Nicholas Lourie,
  Christof Monz, Makoto Morishita, Masaaki Nagata, Ajay Nagesh, Toshiaki
  Nakazawa, Matteo Negri, Santanu Pal, Allahsera~Auguste Tapo, Marco Turchi,
  Valentin Vydrin, and Marcos Zampieri. 2021.
\newblock \href {https://aclanthology.org/2021.wmt-1.1} {Findings of the 2021
  conference on machine translation ({WMT}21)}.
\newblock In \emph{Proc.\ of WMT}.

\bibitem[{Anderson et~al.(2016)Anderson, Fernando, Johnson, and
  Gould}]{SPICE16}
Peter Anderson, Basura Fernando, Mark Johnson, and Stephen Gould. 2016.
\newblock \href {https://arxiv.org/abs/1607.08822} {{SPICE:} semantic
  propositional image caption evaluation}.
\newblock In \emph{Proc.\ of ECCV}.

\bibitem[{Anderson et~al.(2018)Anderson, He, Buehler, Teney, Johnson, Gould,
  and Zhang}]{Anderson2017up-down}
Peter Anderson, Xiaodong He, Chris Buehler, Damien Teney, Mark Johnson, Stephen
  Gould, and Lei Zhang. 2018.
\newblock \href {https://arxiv.org/abs/1707.07998} {Bottom-up and top-down
  attention for image captioning and visual question answering}.
\newblock In \emph{Proc.\ of CVPR}.

\bibitem[{Biten et~al.(2019)Biten, G{\'{o}}mez, Rusi{\~{n}}ol, and
  Karatzas}]{Biten19}
Ali~Furkan Biten, Llu{\'{\i}}s G{\'{o}}mez, Mar{\c{c}}al Rusi{\~{n}}ol, and
  Dimosthenis Karatzas. 2019.
\newblock \href {https://arxiv.org/abs/1904.01475} {Good news, everyone!
  context driven entity-aware captioning for news images}.
\newblock In \emph{Proc.\ of CVPR}.

\bibitem[{Callison-Burch et~al.(2006)Callison-Burch, Osborne, and
  Koehn}]{callison-burch-etal-2006-evaluating}
Chris Callison-Burch, Miles Osborne, and Philipp Koehn. 2006.
\newblock \href {https://www.aclweb.org/anthology/E06-1032} {Re-evaluating the
  role of {BLEU} in machine translation research}.
\newblock In \emph{Proc.\ of EACL}.

\bibitem[{Chen et~al.(2015)Chen, Fang, Lin, Vedantam, Gupta, Doll{\'{a}}r, and
  Zitnick}]{mscoco-captioning}
Xinlei Chen, Hao Fang, Tsung{-}Yi Lin, Ramakrishna Vedantam, Saurabh Gupta,
  Piotr Doll{\'{a}}r, and C.~Lawrence Zitnick. 2015.
\newblock \href {http://arxiv.org/abs/1504.00325} {Microsoft {COCO} captions:
  Data collection and evaluation server}.

\bibitem[{Clark et~al.(2021)Clark, August, Serrano, Haduong, Gururangan, and
  Smith}]{clark-etal-2021-thats}
Elizabeth Clark, Tal August, Sofia Serrano, Nikita Haduong, Suchin Gururangan,
  and Noah~A. Smith. 2021.
\newblock \href {https://arxiv.org/abs/2107.00061} {All that{'}s {`}human{'} is
  not gold: Evaluating human evaluation of generated text}.
\newblock In \emph{Proc.\ of ACL}.

\bibitem[{Cohen(1960)}]{Cohen1960ACO}
Jacob Cohen. 1960.
\newblock \href {https://journals.sagepub.com/doi/10.1177/001316446002000104}
  {A coefficient of agreement for nominal scales}.
\newblock \emph{Educational and Psychological Measurement}.

\bibitem[{Dang(2006)}]{Dang2006OverviewOD}
Hoa~Trang Dang. 2006.
\newblock \href
  {https://www-nlpir.nist.gov/projects/duc/pubs/2006papers/duc2006.pdf}
  {Overview of {DUC} 2006}.
\newblock In \emph{Proc.\ of DUC}.

\bibitem[{Deng et~al.(2009)Deng, Dong, Socher, Li, Li, and
  Fei-Fei}]{deng2009imagenet}
Jia Deng, Wei Dong, Richard Socher, Li-Jia Li, Kai Li, and Li~Fei-Fei. 2009.
\newblock \href {https://arxiv.org/abs/1409.0575} {{ImageNet}: A large-scale
  hierarchical image database}.
\newblock In \emph{Proc.\ of CVPR}.

\bibitem[{Devlin et~al.(2019)Devlin, Chang, Lee, and
  Toutanova}]{devlins2019bert}
Jacob Devlin, Ming-Wei Chang, Kenton Lee, and Kristina Toutanova. 2019.
\newblock \href {https://arxiv.org/abs/810.04805} {{BERT}: Pre-training of deep
  bidirectional transformers for language understanding}.
\newblock In \emph{Proc. of NAACL}.

\bibitem[{Edunov et~al.(2020)Edunov, Ott, Ranzato, and Auli}]{Edunov2020OnTE}
Sergey Edunov, Myle Ott, Marc'Aurelio Ranzato, and Michael Auli. 2020.
\newblock \href {https://arxiv.org/abs/1908.05204} {On the evaluation of
  machine translation systems trained with back-translation}.
\newblock In \emph{Proc.\ of ACL}.

\bibitem[{Elliott and Keller(2013)}]{elliott-keller-2013-image}
Desmond Elliott and Frank Keller. 2013.
\newblock \href {https://aclanthology.org/D13-1128} {Image description using
  visual dependency representations}.
\newblock In \emph{Proc.\ of EMNLP}.

\bibitem[{Elliott and Keller(2014)}]{elliott-keller-2014-comparing}
Desmond Elliott and Frank Keller. 2014.
\newblock \href {https://aclanthology.org/P14-2074} {Comparing automatic
  evaluation measures for image description}.
\newblock In \emph{Proc.\ of ACL}.

\bibitem[{Fabbri et~al.(2021)Fabbri, Kry{\'s}ci{\'n}ski, McCann, Xiong, Socher,
  and Radev}]{fabbri2021summeval}
Alexander~R Fabbri, Wojciech Kry{\'s}ci{\'n}ski, Bryan McCann, Caiming Xiong,
  Richard Socher, and Dragomir Radev. 2021.
\newblock \href {https://arxiv.org/abs/2007.12626} {{SummEval}: Re-evaluating
  summarization evaluation}.
\newblock \emph{TACL}.

\bibitem[{Feinglass and Yang(2021)}]{feinglass-yang-2021-smurf}
Joshua Feinglass and Yezhou Yang. 2021.
\newblock \href {https://arxiv.org/abs/2106.01444} {{SMURF}: {S}e{M}antic and
  linguistic {U}nde{R}standing fusion for caption evaluation via typicality
  analysis}.
\newblock In \emph{Proc.\ of ACL}.

\bibitem[{Freitag et~al.(2021)Freitag, Foster, Grangier, Ratnakar, Tan, and
  Macherey}]{freitag2021experts}
Markus Freitag, George Foster, David Grangier, Viresh Ratnakar, Qijun Tan, and
  Wolfgang Macherey. 2021.
\newblock \href {https://arxiv.org/abs/2104.14478} {Experts, errors, and
  context: A large-scale study of human evaluation for machine translation}.
\newblock \emph{TACL}.

\bibitem[{Gleason et~al.(2019)Gleason, Carrington, Cassidy, Morris, Kitani, and
  Bigham}]{Gleason19}
Cole Gleason, Patrick Carrington, Cameron~Tyler Cassidy, Meredith~Ringel
  Morris, Kris~M. Kitani, and Jeffrey~P. Bigham. 2019.
\newblock \href {https://doi.org/10.1145/3308558.3313605} {"{I}t's almost like
  they're trying to hide it": How user-provided image descriptions have failed
  to make {Twitter} accessible}.
\newblock In \emph{Proc.\ of WWW}.

\bibitem[{Gleason et~al.(2020)Gleason, Pavel, McCamey, Low, Carrington, Kitani,
  and Bigham}]{GleasonCHI20}
Cole Gleason, Amy Pavel, Emma McCamey, Christina Low, Patrick Carrington,
  Kris~M. Kitani, and Jeffrey~P. Bigham. 2020.
\newblock \href
  {https://www.cs.cmu.edu/~jbigham/pubs/pdfs/2020/twitter-a11y.pdf} {Twitter
  a11y: {A} browser extension to make twitter images accessible}.
\newblock In \emph{Proc.\ of CHI}.

\bibitem[{Graham et~al.(2013)Graham, Baldwin, Moffat, and
  Zobel}]{graham-etal-2013-continuous}
Yvette Graham, Timothy Baldwin, Alistair Moffat, and Justin Zobel. 2013.
\newblock \href {https://aclanthology.org/W13-2305} {Continuous measurement
  scales in human evaluation of machine translation}.
\newblock In \emph{Proc.\ of LAW}.

\bibitem[{Graham et~al.(2014)Graham, Baldwin, Moffat, and
  Zobel}]{graham-etal-2014-machine}
Yvette Graham, Timothy Baldwin, Alistair Moffat, and Justin Zobel. 2014.
\newblock \href {https://aclanthology.org/E14-1047} {Is machine translation
  getting better over time?}
\newblock In \emph{Proc.\ of EACL}.

\bibitem[{Graham et~al.(2017)Graham, Baldwin, Moffat, and Zobel}]{GrahamBMZ17}
Yvette Graham, Timothy Baldwin, Alistair Moffat, and Justin Zobel. 2017.
\newblock \href
  {https://www.cambridge.org/core/journals/natural-language-engineering/article/can-machine-translation-systems-be-evaluated-by-the-crowd-alone/E29DA2BC8E6B99AA1481CC92FAB58462}
  {Can machine translation systems be evaluated by the crowd alone}.
\newblock \emph{Natural Language Engineering}.

\bibitem[{Hessel et~al.(2021)Hessel, Holtzman, Forbes, Bras, and
  Choi}]{clipscore}
Jack Hessel, Ari Holtzman, Maxwell Forbes, Ronan~Le Bras, and Yejin Choi. 2021.
\newblock \href {https://arxiv.org/abs/2104.08718} {{CLIPScore}: {A}
  reference-free evaluation metric for image captioning}.
\newblock In \emph{Proc.\ of EMNLP}.

\bibitem[{Hochreiter and Schmidhuber(1997)}]{hochreiter1997long}
Sepp Hochreiter and J{\"u}rgen Schmidhuber. 1997.
\newblock \href {https://dl.acm.org/doi/10.1162/neco.1997.9.8.1735} {Long
  short-term memory}.
\newblock \emph{Neural Computation}.

\bibitem[{Hodosh et~al.(2013)Hodosh, Young, and
  Hockenmaier}]{flicker8k-expert-crowd}
Micah Hodosh, Peter Young, and Julia Hockenmaier. 2013.
\newblock \href {https://www.jair.org/index.php/jair/article/view/10833}
  {Framing image description as a ranking task: Data, models and evaluation
  metrics}.
\newblock \emph{JAIR}.

\bibitem[{Jonsson and Svingby(2007)}]{Jonsson2007TheUO}
Anders Jonsson and Gunilla Svingby. 2007.
\newblock \href
  {https://www.sciencedirect.com/science/article/abs/pii/S1747938X07000188}
  {The use of scoring rubrics: Reliability, validity, and educational
  consequences}.
\newblock \emph{Educational Research Review}.

\bibitem[{Karpathy and Fei-Fei(2015)}]{Karpathy_2015_CVPR}
Andrej Karpathy and Li~Fei-Fei. 2015.
\newblock \href {https://arxiv.org/abs/1412.2306} {Deep visual-semantic
  alignments for generating image descriptions}.
\newblock In \emph{Proc.\ of CVPR}.

\bibitem[{Kasai et~al.(2022)Kasai, Sakaguchi, Bras, Dunagan, Morrison, Fabbri,
  Choi, and Smith}]{billboard}
Jungo Kasai, Keisuke Sakaguchi, Ronan~Le Bras, Lavinia Dunagan, Jacob Morrison,
  Alexander~R. Fabbri, Yejin Choi, and Noah~A. Smith. 2022.
\newblock \href {https://arxiv.org/abs/2112.04139} {Bidimensional leaderboards:
  Generate and evaluate language hand in hand}.
\newblock In \emph{Proc.\ of NAACL}.

\bibitem[{Khashabi et~al.(2021)Khashabi, Stanovsky, Bragg, Lourie, Kasai, Choi,
  Smith, and Weld}]{genie}
Daniel Khashabi, Gabriel Stanovsky, Jonathan Bragg, Nicholas Lourie, Jungo
  Kasai, Yejin Choi, Noah~A. Smith, and Daniel~S. Weld. 2021.
\newblock \href {https://arxiv.org/abs/2101.06561} {{GENIE:} {A} leaderboard
  for human-in-the-loop evaluation of text generation}.

\bibitem[{Kilickaya et~al.(2017)Kilickaya, Erdem, Ikizler-Cinbis, and
  Erdem}]{kilickaya-etal-2017-evaluating}
Mert Kilickaya, Aykut Erdem, Nazli Ikizler-Cinbis, and Erkut Erdem. 2017.
\newblock \href {https://arxiv.org/abs/1612.07600} {Re-evaluating automatic
  metrics for image captioning}.
\newblock In \emph{Proc.\ of EACL}.

\bibitem[{Koehn(2004)}]{koehn-2004-statistical}
Philipp Koehn. 2004.
\newblock \href {https://aclanthology.org/W04-3250} {Statistical significance
  tests for machine translation evaluation}.
\newblock In \emph{Proc.\ of EMNLP}.

\bibitem[{Krishna et~al.(2016)Krishna, Zhu, Groth, Johnson, Hata, Kravitz,
  Chen, Kalanditis, Li, Shamma, Bernstein, and Fei-Fei}]{krishnavisualgenome}
Ranjay Krishna, Yuke Zhu, Oliver Groth, Justin Johnson, Kenji Hata, Joshua
  Kravitz, Stephanie Chen, Yannis Kalanditis, Li-Jia Li, David~A Shamma,
  Michael Bernstein, and Li~Fei-Fei. 2016.
\newblock \href {https://arxiv.org/abs/1602.07332} {{Visual Genome}: Connecting
  language and vision using crowdsourced dense image annotations}.
\newblock \emph{IJCV}.

\bibitem[{Kryscinski et~al.(2019)Kryscinski, Keskar, McCann, Xiong, and
  Socher}]{kryscinski-etal-2019-neural}
Wojciech Kryscinski, Nitish~Shirish Keskar, Bryan McCann, Caiming Xiong, and
  Richard Socher. 2019.
\newblock \href {https://arxiv.org/abs/1908.08960} {Neural text summarization:
  A critical evaluation}.
\newblock In \emph{Proc.\ of EMNLP}.

\bibitem[{Kulkarni et~al.(2011)Kulkarni, Premraj, Dhar, Li, Choi, Berg, and
  Berg}]{babytalk2011}
Girish Kulkarni, Visruth Premraj, Sagnik Dhar, Siming Li, Yejin Choi,
  Alexander~C Berg, and Tamara~L Berg. 2011.
\newblock \href {http://www.tamaraberg.com/papers/generation_cvpr11.pdf} {Baby
  talk: Understanding and generating simple image descriptions}.
\newblock In \emph{Proc.\ of CVPR}.

\bibitem[{Li et~al.(2011)Li, Kulkarni, Berg, Berg, and
  Choi}]{li-etal-2011-composing}
Siming Li, Girish Kulkarni, Tamara~L Berg, Alexander~C Berg, and Yejin Choi.
  2011.
\newblock \href {https://aclanthology.org/W11-0326} {Composing simple image
  descriptions using web-scale n-grams}.
\newblock In \emph{Proc.\ of CoNLL}.

\bibitem[{Li et~al.(2020)Li, Yin, Li, Hu, Zhang, Zhang, Wang, Hu, Dong, Wei,
  Choi, and Gao}]{li2020oscar}
Xiujun Li, Xi~Yin, Chunyuan Li, Xiaowei Hu, Pengchuan Zhang, Lei Zhang, Lijuan
  Wang, Houdong Hu, Li~Dong, Furu Wei, Yejin Choi, and Jianfeng Gao. 2020.
\newblock \href {https://arxiv.org/abs/2004.06165} {Oscar: Object-semantics
  aligned pre-training for vision-language tasks}.
\newblock In \emph{Proc.\ of ECCV}.

\bibitem[{Lin(2004)}]{Lin2004ROUGEAP}
Chin-Yew Lin. 2004.
\newblock \href {https://www.aclweb.org/anthology/W04-1013/} {{ROUGE}: A
  package for automatic evaluation of summaries}.
\newblock In \emph{Proc.\ of Text Summarization Branches Out}.

\bibitem[{Lin et~al.(2014)Lin, Maire, Belongie, Bourdev, Girshick, Hays,
  Perona, Ramanan, Doll{\'{a}}r, and Zitnick}]{mscoco}
Tsung{-}Yi Lin, Michael Maire, Serge~J. Belongie, Lubomir~D. Bourdev, Ross~B.
  Girshick, James Hays, Pietro Perona, Deva Ramanan, Piotr Doll{\'{a}}r, and
  C.~Lawrence Zitnick. 2014.
\newblock \href {http://arxiv.org/abs/1405.0312} {Microsoft {COCO:} common
  objects in context}.
\newblock In \emph{Proc.\ of ECCV}.

\bibitem[{Ma et~al.(2019)Ma, Wei, Bojar, and Graham}]{ma-etal-2019-results}
Qingsong Ma, Johnny Wei, Ond{\v{r}}ej Bojar, and Yvette Graham. 2019.
\newblock \href {https://www.aclweb.org/anthology/W19-5302} {Results of the
  {WMT}19 metrics shared task: Segment-level and strong {MT} systems pose big
  challenges}.
\newblock In \emph{Proc.\ of WMT}.

\bibitem[{Miller(1995)}]{miller1995wordnet}
George~A. Miller. 1995.
\newblock \href {https://mitpress.mit.edu/books/wordnet} {{WordNet}: A lexical
  database for {E}nglish}.
\newblock \emph{CACM}.

\bibitem[{Misra et~al.(2016)Misra, Zitnick, Mitchell, and
  Girshick}]{Misra2016SeeingTT}
Ishan Misra, C.~Lawrence Zitnick, Margaret Mitchell, and Ross~B. Girshick.
  2016.
\newblock \href {https://arxiv.org/abs/1512.06974} {Seeing through the human
  reporting bias: Visual classifiers from noisy human-centric labels}.
\newblock In \emph{Proc.\ of CVPR}.

\bibitem[{Mitchell et~al.(2012)Mitchell, Dodge, Goyal, Yamaguchi, Stratos, Han,
  Mensch, Berg, Berg, and Daum{\'e}~III}]{mitchell-etal-2012-midge}
Margaret Mitchell, Jesse Dodge, Amit Goyal, Kota Yamaguchi, Karl Stratos,
  Xufeng Han, Alyssa Mensch, Alex Berg, Tamara Berg, and Hal Daum{\'e}~III.
  2012.
\newblock \href {https://aclanthology.org/E12-1076} {{M}idge: Generating image
  descriptions from computer vision detections}.
\newblock In \emph{Proc.\ of EACL}.

\bibitem[{Ordonez et~al.(2013)Ordonez, Deng, Choi, Berg, and
  Berg}]{entry-level-categories}
Vicente Ordonez, Jia Deng, Yejin Choi, Alexander~C. Berg, and Tamara~L. Berg.
  2013.
\newblock \href {https://ieeexplore.ieee.org/document/6751455} {From large
  scale image categorization to entry-level categories}.
\newblock In \emph{Proc.\ of ICCV}.

\bibitem[{Papineni et~al.(2002)Papineni, Roukos, Ward, and
  Zhu}]{Papineni2001BleuAM}
Kishore Papineni, Salim Roukos, Todd Ward, and Wei-Jing Zhu. 2002.
\newblock \href {https://www.aclweb.org/anthology/P02-1040.pdf} {{BLEU}: a
  method for automatic evaluation of machine translation}.
\newblock In \emph{Proc.\ of ACL}.

\bibitem[{Post(2018)}]{post-2018-call}
Matt Post. 2018.
\newblock \href {https://www.aclweb.org/anthology/W18-6319} {A call for clarity
  in reporting {BLEU} scores}.
\newblock In \emph{Proc.\ of WMT}.

\bibitem[{Radford et~al.(2021)Radford, Kim, Hallacy, Ramesh, Goh, Agarwal,
  Sastry, Askell, Mishkin, Clark, Krueger, and Sutskever}]{clip2021}
Alec Radford, Jong~Wook Kim, Chris Hallacy, Aditya Ramesh, Gabriel Goh,
  Sandhini Agarwal, Girish Sastry, Amanda Askell, Pamela Mishkin, Jack Clark,
  Gretchen Krueger, and Ilya Sutskever. 2021.
\newblock \href {https://arxiv.org/abs/2103.00020} {Learning transferable
  visual models from natural language supervision}.

\bibitem[{Rashtchian et~al.(2010)Rashtchian, Young, Hodosh, and
  Hockenmaier}]{rashtchian-etal-2010-collecting}
Cyrus Rashtchian, Peter Young, Micah Hodosh, and Julia Hockenmaier. 2010.
\newblock \href {https://aclanthology.org/W10-0721} {Collecting image
  annotations using {A}mazon{'}s {M}echanical {T}urk}.
\newblock In \emph{Proc.\ of NAACL Workshop on Creating Speech and Language
  Data with {A}mazon{'}s Mechanical Turk}.

\bibitem[{Ren et~al.(2015)Ren, He, Girshick, and Sun}]{fasterRCNN}
Shaoqing Ren, Kaiming He, Ross Girshick, and Jian Sun. 2015.
\newblock \href {https://arxiv.org/abs/1506.01497} {Faster {R-CNN}: Towards
  real-time object detection with region proposal networks}.
\newblock In \emph{Proc.\ of NeurIPS}.

\bibitem[{Rohrbach et~al.(2018)Rohrbach, Hendricks, Burns, Darrell, and
  Saenko}]{rohrbach-etal-2018-object}
Anna Rohrbach, Lisa~Anne Hendricks, Kaylee Burns, Trevor Darrell, and Kate
  Saenko. 2018.
\newblock \href {https://arxiv.org/abs/1809.02156} {Object hallucination in
  image captioning}.
\newblock In \emph{Proc.\ of EMNLP}.

\bibitem[{Sakaguchi et~al.(2017)Sakaguchi, Post, and
  Van~Durme}]{sakaguchi-etal-2017-grammatical}
Keisuke Sakaguchi, Matt Post, and Benjamin Van~Durme. 2017.
\newblock \href {https://aclanthology.org/I17-2062} {Grammatical error
  correction with neural reinforcement learning}.
\newblock In \emph{Proc.\ of IJCNLP}.

\bibitem[{Schuster et~al.(2015)Schuster, Krishna, Chang, Fei-Fei, and
  Manning}]{schuster-etal-2015-generating}
Sebastian Schuster, Ranjay Krishna, Angel Chang, Li~Fei-Fei, and Christopher~D.
  Manning. 2015.
\newblock \href {https://aclanthology.org/W15-2812} {Generating semantically
  precise scene graphs from textual descriptions for improved image retrieval}.
\newblock In \emph{Proc.\ of VL}.

\bibitem[{Sharma et~al.(2018)Sharma, Ding, Goodman, and
  Soricut}]{sharma-etal-2018-conceptual}
Piyush Sharma, Nan Ding, Sebastian Goodman, and Radu Soricut. 2018.
\newblock \href {https://aclanthology.org/P18-1238} {{C}onceptual {C}aptions: A
  cleaned, hypernymed, image alt-text dataset for automatic image captioning}.
\newblock In \emph{Proc.\ of ACL}.

\bibitem[{Shuster et~al.(2019)Shuster, Humeau, Hu, Bordes, and
  Weston}]{Shuster2019}
Kurt Shuster, Samuel Humeau, Hexiang Hu, Antoine Bordes, and Jason Weston.
  2019.
\newblock \href {https://arxiv.org/abs/1810.10665} {Engaging image captioning
  via personality}.
\newblock In \emph{Proc.\ of CVPR}.

\bibitem[{van Miltenburg(2020)}]{van-miltenburg-2020-image}
Emiel van Miltenburg. 2020.
\newblock \href {https://aclanthology.org/2020.lantern-1.4} {How do image
  description systems describe people? a targeted assessment of system
  competence in the {PEOPLE}-domain}.
\newblock In \emph{Proc.\ of LANTERN}.

\bibitem[{van Miltenburg et~al.(2017)van Miltenburg, Elliott, and
  Vossen}]{MiltenburgEV17}
Emiel van Miltenburg, Desmond Elliott, and Piek Vossen. 2017.
\newblock \href {https://arxiv.org/abs/1707.01736} {Cross-linguistic
  differences and similarities in image descriptions}.
\newblock In \emph{Proc.\ of INLG}.

\bibitem[{van Miltenburg et~al.(2018)van Miltenburg, Elliott, and
  Vossen}]{van-miltenburg-etal-2018-measuring}
Emiel van Miltenburg, Desmond Elliott, and Piek Vossen. 2018.
\newblock \href {https://aclanthology.org/C18-1147/} {Measuring the diversity
  of automatic image descriptions}.
\newblock In \emph{Proc.\ of COLING}.

\bibitem[{Vaswani et~al.(2017)Vaswani, Shazeer, Parmar, Uszkoreit, Jones,
  Gomez, Kaiser, and Polosukhin}]{Vaswani2017AttentionIA}
Ashish Vaswani, Noam Shazeer, Niki Parmar, Jakob Uszkoreit, Llion Jones,
  Aidan~N. Gomez, \L{}ukasz Kaiser, and Illia Polosukhin. 2017.
\newblock \href {https://arxiv.org/pdf/1706.03762.pdf} {Attention is all you
  need}.
\newblock In \emph{Proc. of NeurIPS}.

\bibitem[{Vedantam et~al.(2015)Vedantam, Zitnick, and Parikh}]{cider2015}
Ramakrishna Vedantam, C.~Lawrence Zitnick, and Devi Parikh. 2015.
\newblock \href {https://arxiv.org/abs/1411.5726} {{CIDEr}: Consensus-based
  image description evaluation}.
\newblock In \emph{Proc.\ of CVPR}.

\bibitem[{Wang et~al.(2020)Wang, Feng, Narasimhan, and
  Russakovsky}]{wang2020spiceu}
Zeyu Wang, Berthy Feng, Karthik Narasimhan, and Olga Russakovsky. 2020.
\newblock \href {https://arxiv.org/abs/2009.03949} {Towards unique and
  informative captioning of images}.
\newblock In \emph{Proc.\ of ECCV}.

\bibitem[{Wolf et~al.(2020)Wolf, Debut, Sanh, Chaumond, Delangue, Moi, Cistac,
  Rault, Louf, Funtowicz, Davison, Shleifer, von Platen, Ma, Jernite, Plu, Xu,
  Le~Scao, Gugger, Drame, Lhoest, and Rush}]{wolf-etal-2020-transformers}
Thomas Wolf, Lysandre Debut, Victor Sanh, Julien Chaumond, Clement Delangue,
  Anthony Moi, Pierric Cistac, Tim Rault, Remi Louf, Morgan Funtowicz, Joe
  Davison, Sam Shleifer, Patrick von Platen, Clara Ma, Yacine Jernite, Julien
  Plu, Canwen Xu, Teven Le~Scao, Sylvain Gugger, Mariama Drame, Quentin Lhoest,
  and Alexander Rush. 2020.
\newblock \href {https://aclanthology.org/2020.emnlp-demos.6} {{HuggingFace}'s
  transformers: State-of-the-art natural language processing}.
\newblock In \emph{Proc.\ of EMNLP: System Demonstrations}.

\bibitem[{Xie et~al.(2017)Xie, Girshick, Doll{\'{a}}r, Tu, and He}]{C4}
Saining Xie, Ross~B. Girshick, Piotr Doll{\'{a}}r, Zhuowen Tu, and Kaiming He.
  2017.
\newblock \href {https://arxiv.org/abs/1611.05431} {Aggregated residual
  transformations for deep neural networks}.
\newblock In \emph{Proc.\ of CVPR}.

\bibitem[{Yang et~al.(2011)Yang, Teo, Daum{\'e}~III, and
  Aloimonos}]{yang-etal-2011-corpus}
Yezhou Yang, Ching Teo, Hal Daum{\'e}~III, and Yiannis Aloimonos. 2011.
\newblock \href {https://aclanthology.org/D11-1041} {Corpus-guided sentence
  generation of natural images}.
\newblock In \emph{Proc.\ of EMNLP}.

\bibitem[{Yatskar et~al.(2014)Yatskar, Galley, Vanderwende, and
  Zettlemoyer}]{yatskar-etal-2014-see}
Mark Yatskar, Michel Galley, Lucy Vanderwende, and Luke Zettlemoyer. 2014.
\newblock \href {https://aclanthology.org/S14-1015} {See no evil, say no evil:
  Description generation from densely labeled images}.
\newblock In \emph{Proc.\ of *SEM}.

\bibitem[{Young et~al.(2014)Young, Lai, Hodosh, and Hockenmaier}]{flicker30k}
Peter Young, Alice Lai, Micah Hodosh, and Julia Hockenmaier. 2014.
\newblock \href {https://aclanthology.org/Q14-1006} {From image descriptions to
  visual denotations: New similarity metrics for semantic inference over event
  descriptions}.
\newblock \emph{TACL}.

\bibitem[{Yuan and Briscoe(2016)}]{yuan-briscoe-2016-grammatical}
Zheng Yuan and Ted Briscoe. 2016.
\newblock \href {https://aclanthology.org/N16-1042} {Grammatical error
  correction using neural machine translation}.
\newblock In \emph{Proc.\ of NAACL}.

\bibitem[{Zhang et~al.(2021)Zhang, Li, Hu, Yang, Zhang, Wang, Choi, and
  Gao}]{zhang2021vinvl}
Pengchuan Zhang, Xiujun Li, Xiaowei Hu, Jianwei Yang, Lei Zhang, Lijuan Wang,
  Yejin Choi, and Jianfeng Gao. 2021.
\newblock \href {https://arxiv.org/abs/2101.00529} {{VinVL}: Making visual
  representations matter in vision-language models}.
\newblock In \emph{Proc.\ of CVPR 2021}.

\bibitem[{Zhang et~al.(2020)Zhang, Kishore, Wu, Weinberger, and
  Artzi}]{bertscore}
Tianyi Zhang, Varsha Kishore, Felix Wu, Kilian~Q. Weinberger, and Yoav Artzi.
  2020.
\newblock \href {https://arxiv.org/abs/1904.09675} {{BERTScore}: Evaluating
  text generation with {BERT}}.
\newblock In \emph{Proc.\ of ICLR}.

\bibitem[{Zhou et~al.(2020)Zhou, Palangi, Zhang, Hu, Corso, and
  Gao}]{Unified-VLP}
Luowei Zhou, Hamid Palangi, Lei Zhang, Houdong Hu, Jason~J. Corso, and Jianfeng
  Gao. 2020.
\newblock \href {https://arxiv.org/abs/1909.11059} {Unified vision-language
  pre-training for image captioning and {VQA}}.
\newblock In \emph{Proc.\ of AAAI}.

\end{thebibliography}
\bibliographystyle{acl_natbib}

\newpage
\appendix
\newpage
\section{Appendix}
\subsection{Fluency Rubrics} \label{sec:fluencyrubrics}
Table \ref{tab:fluency_rubrics} presents our fluency rubrics.
They were developed by the first four authors (two of whom were native English speakers, and one was a graduate student in linguistics).
Generally, if a fluency problem is expected to be easily corrected by a text postprocessing algorithm, the penalty is 0.1. More severe errors (e.g., broken sentence and ambiguity) are penalized more.

\begin{table*}[h]
\small
\centering
\addtolength{\tabcolsep}{-2.6pt}  
\def\arraystretch{1.2}
\begin{tabular}{@{} lcl @{}}
\toprule
\textbf{Fluency Error Type}& Penalty & Example \\
\hline
Obvious spelling error, one vs.\ two words & 0.1 & \textit{cel phone}, \textit{surf board} \\
Grammatical error that can be easily fixed & 0.1 & \textit{a otter}\\
Casing issue & 0.1 & \textit{tv}, \textit{christmas} \\
Hyphenation & 0.1 & \textit{horse drawn carriage} \\
Interpretable but unnatural wording & 0.1 & \textit{double decked bus} \\
Non-trivial punctuation & 0.2 & \textit{A bird standing in the wooded area with leaves all around. } \\
Misleading spelling error & 0.5 &\textit{A good stands in the grass next to the water.} (good$\rightarrow$goose) \\
Duplication & 0.5 & \textit{A display case of donuts and doughnuts.} \\
Ambiguity & 0.5 & \textit{A cat is on a table with a cloth on it.} \\
Awkward construction & 0.1--0.5 & \textit{There is a freshly made pizza out of the oven.}\\
Broken sentence & 0.5+ & \textit{A large concrete sign small buildings behind it.}  \\
\bottomrule
\end{tabular}
\caption{Fluency penalty rubrics.}
\label{tab:fluency_rubrics}
\end{table*}

\subsection{Automatic Metrics}
\label{sec:decription_metrics}
Here we discuss details and configurations of the automatic metrics used in \S\ref{sec:metric_comparisons}.
CLIPScore and RefCLPScore use image features from CLIP \cite{clip2021}, a crossmodal retrieval model trained on 400M image-caption pairs from the web.
All the other five metrics only use reference captions.
\paragraph{BLEU}
BLEU \cite{Papineni2001BleuAM} is a precision-oriented metric and measures n-gram overlap between the generated and reference captions.
We use the \textsc{SacreBLEU} implementation of BLEU-4 and get sentence-level scores \cite{post-2018-call}.\footnote{\url{https://github.com/mjpost/sacreBLEU/blob/v1.2.12/sacrebleu.py\#L999}.}
\paragraph{ROUGE}
ROUGE \cite{Lin2004ROUGEAP} measures the number of overlapping n-grams between the generated and reference captions. We use the rouge-score implementation of ROUGE-L.\footnote{\url{https://pypi.org/project/rouge-score/}.}
\paragraph{CIDEr} CIDEr \cite{cider2015} measures the cosine similarity between the n-gram counts of the generated and reference captions with TF-IDF weighting.
We use the implementation from the pycocoevalcap package.\footnote{\url{https://github.com/salaniz/pycocoevalcap}.}
\paragraph{SPICE} SPICE \cite{SPICE16} predicts scene graphs from the generated and reference captions using the Stanford scene graph parser \cite{schuster-etal-2015-generating}.
It then measures the $F_1$ score between scene graphs from the generated and reference captions.
WordNet Synsets are used to cluster synonyms \cite{miller1995wordnet}.
We again use the implementation from the pycocoevalcap package.
\paragraph{BERTScore}
BERTScore \cite{bertscore} aligns tokens between the generated and reference captions using contextual word representations from BERT \cite{devlins2019bert}.
We use the HuggingFace implementation \cite{wolf-etal-2020-transformers} and compute the $F_1$ score.
As in \citet{bertscore}, we take the maximum score over all reference captions.
\paragraph{CLIPScore}
CLIPScore \cite{clipscore} is the only \textit{referenceless} metric out of the 7 metrics.
It measures the cosine similarity between the generated caption and given image using the representations from CLIP.
It is shown to correlate better with human judgments from prior work, compared to previous reference-based metrics \cite{clipscore}.
We use the official implementation by the authors.\footnote{\url{https://github.com/jmhessel/pycocoevalcap}.}
\paragraph{RefCLIPScore}
RefCLIPScore augments CLIPScore with the maximum similarity between the generated and reference captions.
We again use the official implementation.

\subsection{Evaluated Captions}
\label{app:evaluated-captions}
We evaluated the following four strong models from the literature as well as human-generated captions.
They share similar pipeline structure but vary in model architecture, (pre)training data, model size, and (pre)training objective. Evaluating captions from them will enable us to better understand what has been improved and what is still left to future captioning models.
\paragraph{Up-Down}
The bottom-up and top-down attention model (Up-Down, \citealp{Anderson2017up-down}) performs pipelined image captioning: \textit{object detection} that finds objects and their corresponding image regions and \textit{crossmodal generation} that predicts a caption based on the features from object detection.
The bottom-up attention finds salient image regions during object detection, and the top-down one attends to these regions during crossmodal generation.
Up-Down uses Faster R-CNN \cite{fasterRCNN} and LSTMs \cite{hochreiter1997long} for object detection and crossmodal generation respectively.
Faster R-CNN is trained with the Visual Genome dataset \cite{krishnavisualgenome}, and the crossmodal generation model is trained on the MSCOCO dataset.
We generate captions for the test data with a model optimized with crossentropy.\footnote{\url{https://vision-explorer.allenai.org/image_captioning}.}

\paragraph{Unified-VLP}
Unified-VLP \cite{Unified-VLP} also runs a pipeline of object detection and crossmodal generation.
Faster R-CNN and the transformer architecture \cite{Vaswani2017AttentionIA} are used for object detection and crossmodal generation respectively.
Similar to Up-Down, the Faster R-CNN object detection model is trained with the Visual Genome dataset.
The transformer generation model, on the other hand, is initialized with base-sized BERT \cite{devlins2019bert} and pretrained on the Conceptual Captions dataset (3M images, \citealp{sharma-etal-2018-conceptual}) with the masked and left-to-right language modeling objectives for the captions.
The crossmodal generation model is then finetuned on the MSCOCO dataset. 
We apply beam search of size 5 to the model with CIDEr optimization.

\paragraph{VinVL-base, VinVL-large}
VinVL with Oscar \cite{li2020oscar, zhang2021vinvl} performs a similar pipeline of object detection, followed by crossmodal generation.
The crossmodal model is initialized with BERT \cite{devlins2019bert} as in Unified-VLP but uses detected object tags to encourage alignments between image features and word representations. 
The object detection model with the ResNeXt-152 C4 architecture \cite{C4} is pretrained with ImageNet \cite{deng2009imagenet} and trained on 2.5M images from various datasets.
The transformer-based crossmodal generator is initialized with BERT, pretrained with 5.7M images, and finetuned for MSCOCO captioning.
We use VinVL-base and VinVL-large that are both finetuned with CIDEr optimization\footnote{\url{https://github.com/microsoft/Oscar/blob/master/VinVL_MODEL_ZOO.md\#Image-Captioning-on-COCO}.} and generate captions with beam search of size 5.

\paragraph{Human}
In addition to machine-generated captions from the four models, we assessed the quality of human-generated reference captions from MSCOCO.
This will allow us to understand the performance gap between machines and humans, as well as the quality of crowdsourced captions.
Human-generated captions were created using Amazon Mechanical Turk \cite{mscoco-captioning}.
Crowdworkers were only given the following instructions \cite{mscoco-captioning}:
\begin{itemize}[noitemsep, leftmargin=*, topsep=0pt]
\item Describe all the important parts of the scene.
\item Do not start the sentences with ``There is.''
\item Do not describe unimportant details.
\item Do not describe things that might have happened
in the future or past.
\item Do not describe what a person might say.
\item Do not give people proper names.
\item The sentences should contain at least 8 words.
\end{itemize}
Every image has five human-generated captions, and we randomly selected one for each to evaluate.
We found, however, a non-negligible number of noisy captions in the MSCCOCO dataset from annotation spammers.
We often find subjective adjectives (e.g., \textit{very nice/clean/cute}) or words that diverge from \textit{inclusive language} in reference captions, probably because crowdworkers increased the number of words in captions effortlessly (see the last instruction item that says captions have to have 8+ words).
To better estimate the performance of a human that invests reasonable effort into the captioning task, we resampled a caption for 13\% of the test images, which would have been given a total score lower than 4.0.

\subsection{Additional Machine vs.\ Human Examples}
\label{sec:additional_machine_vs_human}
Table \ref{table:additiona_machine_vs_human} provides an additional example that contrasts machine- and human-generated captions.
All machines generate generic captions and ignore the most important information that a traditional Thanksgiving dinner is being served on the table.
\begin{table*}[h]
    \small
     \begin{center}
     \begin{tabular}{ @{}  c p{8.5cm}  c  c c   @{} }
     \toprule
      Image & Caption & P & R & Total \\ 
    \cmidrule(lr){1-1}\cmidrule(lr){2-2}
    \cmidrule(lr){3-4}
    \cmidrule(lr){5-5}
    \raisebox{0.0cm}{
     \multirow{10}{*}{
     {\includegraphics[width=0.21\textwidth]{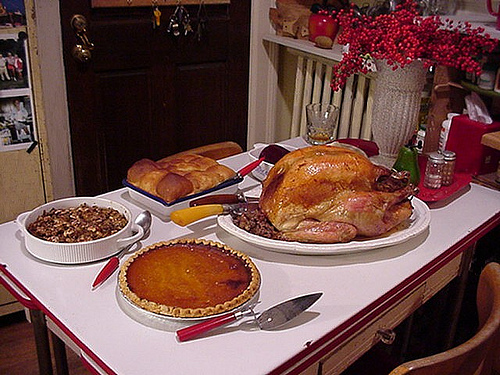}}
     }}
     & 
     \textbf{10-A}: Up-Down
     &  
     \multirow{2}{*}{5}
     &
     \multirow{2}{*}{2}
     &
     \multirow{2}{*}{3.5}
     \\ 
     & 
     \textbf{\textit{A table that has some food on it.}}
     &&&
     \\
     \cmidrule(lr){2-5}
     & 
     \textbf{10-B}: Unified-VLP
     &
     \multirow{2}{*}{5}
     &
     \multirow{2}{*}{2}
     &
     \multirow{2}{*}{3.5}
     \\ 
     &
     \textbf{\textit{A table with plates of food on a table.}}
     &&&
     \\
     
     \cmidrule(lr){2-5}
     & 
     \textbf{10-C}: VinVL-base
     & 
     \multirow{2}{*}{4}
     &
     \multirow{2}{*}{2}
     &
     \multirow{2}{*}{3}
     \\
     & 
     \textbf{\textit{A red table topped with plates of food and bowls of food.}}
     &&&
     \\ 
     
     \cmidrule(lr){2-5}
     & 
     \textbf{10-D}: VinVL-large
     & 
     \multirow{2}{*}{5}
     &
     \multirow{2}{*}{3}
     &
     \multirow{2}{*}{4}
     \\
     & 
     \textbf{\textit{A table with a turkey and other food on it.}}
     &&&
     \\ 
     
     \cmidrule(lr){2-5}
     & 
     \textbf{10-E}: Human
     & 
     \multirow{2}{*}{5}
     &
     \multirow{2}{*}{5}
     &
     \multirow{2}{*}{5}
     \\
     & 
     \textbf{\textit{A table set for a traditional Thanksgiving dinner.}}
     &&&
     \\ 
      \bottomrule
      \end{tabular}
      \caption{Additional example that contrasts machine- and human-generated captions. Similar to Table \ref{table:machine_vs_human}, machine-generated captions ignore the most salient information: Thanksgiving dinner.
      Note that this case is specific to North America; such salient information can vary across cultures or languages \cite{MiltenburgEV17}.
      None of these captions are penalized for fluency, conciseness, or inclusive language.}
      \label{table:additiona_machine_vs_human}
      \end{center}
\end{table*}

\end{document}